\documentclass[a4paper,fleqn]{cas-dc}

\usepackage[numbers]{natbib}
\usepackage{pifont}
\usepackage{makecell}
\usepackage{hyperref}
\usepackage{xcolor}

\def\tsc#1{\csdef{#1}{\textsc{\lowercase{#1}}\xspace}}
\tsc{WGM}
\tsc{QE}

\begin{document}
\let\WriteBookmarks\relax
\def\floatpagepagefraction{1}
\def\textpagefraction{.001}

\shorttitle{Survey: MLLM for Chart Understanding}    

\shortauthors{Z. Yi et al.}

\title [mode = title]{Multimodal Information Fusion for Chart Understanding: A Survey of MLLMs—Evolution, Limitations, and Cognitive Enhancement}  



%

\author[1, 2]{Zhihang Yi}



\ead{zhihangyi0224@gmail.com}

\ead[url]{https://github.com/yzhbradoodrrpurp}

\credit{Conceptualization, Methodology, Investigation, Data Curation, Writing – Original Draft Preparation, Visualization}

\author[3]{Jian Zhao}
\ead{zhaoj90@chinatelecom.cn}
\ead[url]{https://zhaoj9014.github.io}
\credit{Resources, Validation}

\author[1,2]{Jiancheng Lv}
\ead{lvjiancheng@scu.edu.cn}
\ead[url]{https://center.dicalab.cn/}
\credit{Resources, Validation}

\affiliation[1]{organization={College of Computer Science, Sichuan University},
            city={Chengdu},
            postcode={610065}, 
            state={Sichuan},
            country={P.R. China}}

\author[1, 2]{Tao Wang}

\cormark[1]


\ead{twangnh@gmail.com}

\ead[url]{https://twangnh.github.io}

\credit{Supervision, Writing – Review \& Editing, Project Administration}

\affiliation[2]{organization={Engineering Research Center of Machine Learning and Industry Intelligence, Ministry of Education},
            city={Chengdu},
            postcode={610065}, 
            state={Sichuan},
            country={P.R. China}}

\cortext[1]{Corresponding author}



\affiliation[3]{organization={China Telecom Institute of AI},
            city={Beijing},
            postcode={100053},
            country={P. R. China}}

\begin{abstract}
Chart understanding is a quintessential information fusion task, requiring the seamless integration of graphical and textual data to extract meaning. The advent of Multimodal Large Language Models (MLLMs) has revolutionized this domain, yet the landscape of MLLM-based chart analysis remains fragmented and lacks systematic organization. This survey provides a comprehensive roadmap of this nascent frontier by structuring the domain's core components. We begin by analyzing the fundamental challenges of fusing visual and linguistic information in charts. We then categorize downstream tasks and datasets, introducing a novel taxonomy of canonical and non-canonical benchmarks to highlight the field's expanding scope. Subsequently, we present a comprehensive evolution of methodologies, tracing the progression from classic deep learning techniques to state-of-the-art MLLM paradigms that leverage sophisticated fusion strategies. By critically examining the limitations of current models, particularly their perceptual and reasoning deficits, we identify promising future directions, including advanced alignment techniques and reinforcement learning for cognitive enhancement. This survey aims to equip researchers and practitioners with a structured understanding of how MLLMs are transforming chart information fusion and to catalyze progress toward more robust and reliable systems.
\end{abstract}





\maketitle


\section{Introduction}
\subsection{The Emergence of MLLMs}
The development of Multimodal Large Language Models (MLLMs) has progressed enormously in the past few years, profoundly transforming the landscape of Computer Vision. With the emergence of GPT \cite{achiam2023gpt}, Gemini \cite{team2023gemini, team2024gemma}, DeepSeek \cite{guo2024deepseek, liu2024deepseek, zhu2024deepseek, wu2024deepseekvl2mixtureofexpertsvisionlanguagemodels}, Llava \cite{touvron2023llama, touvron2023llama2, liu2023visual}, QWen \cite{bai2025qwen25vltechnicalreport} and so on, MLLMs have demonstrated unprecedented capabilities in understanding and reasoning across multiple modalities, including text, images and even videos. These models leverage powerful Transformer-based architectures \cite{vaswani2017attention, dosovitskiy2020image, liu2021swin} and Visual Instruction-Tuning \cite{liu2023visual} to align language with visual information in a coherent manner. 

The proposal of Vision Transformer \cite{dosovitskiy2020image} and CLIP \cite{radford2021learning} symbolizes the modern vision-language foundation model era. Vision Transformer (ViT) demonstrates how pure Transformer architecture, originally designed for Natural Language Processing tasks, can be applied to visual tasks. CLIP introduces a contrastive learning paradigm that aims to align visual representations with their corresponding textual representations in a shared embedding space. Based on ViT and CLIP, earlier MLLMs, such as Flamingo \cite{alayrac2022flamingo} and BLIP \cite{li2022blip, li2023blip}, made successful attempts at combining visual and textual representations together. They leverage pre-trained Vision Encoders (often ViTs) and Language Models \cite{devlin2019bert, hoffmann2022training, chiang2023vicuna}, connected via Cross-Modal Transformers or Q-Former Layers, to align model outputs with human-like instruction-following behaviors.

Together, these developments have laid a solid foundation for subsequent larger-scale MLLMs, enabling the modern general-purpose MLLMs not only to recognize and describe visual contents, but also to engage in complex reasoning tasks.

\begin{figure*}[htbp]
    \centering
    \includegraphics[width=\linewidth]{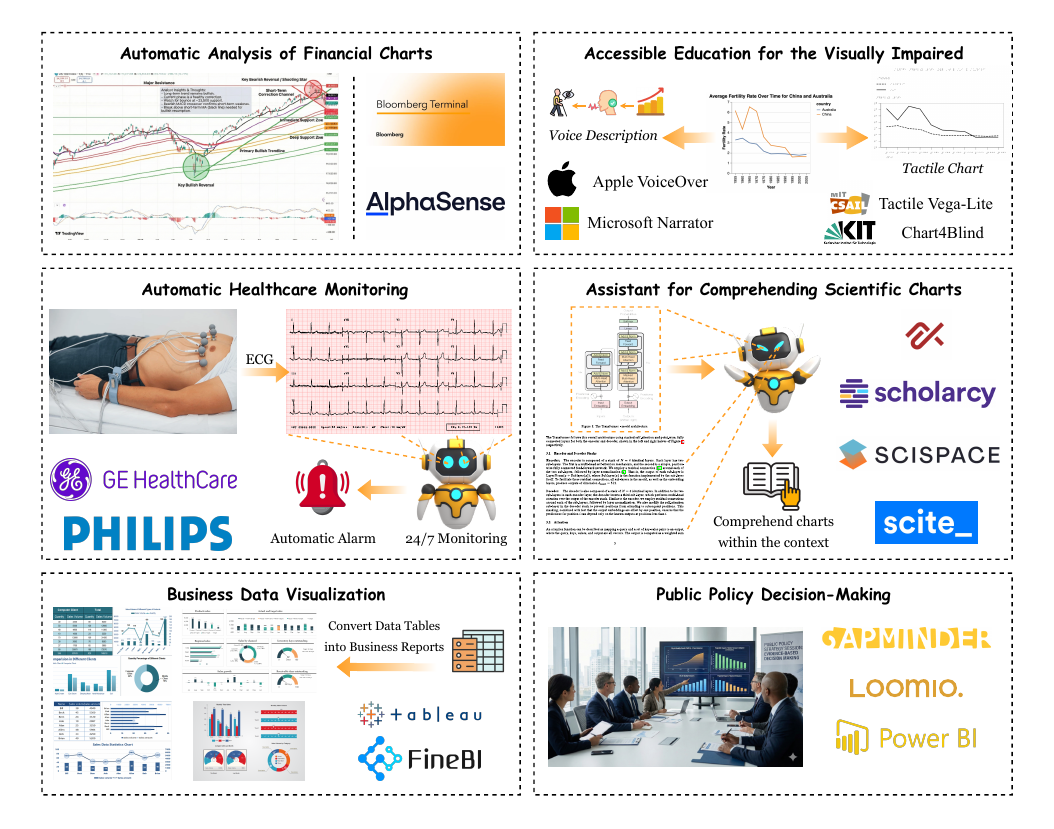}
    \caption{The prevalence and application potential of charts. Chart understanding can support financial analysis, accessibility through voice descriptions or tactile charts \cite{Moured_2024, 2025-tactile-vega-lite}, biomedical signal interpretation, scientific literature reading, business intelligence, and public-policy analysis. These scenarios motivate more reliable models that can fuse visual structure, textual annotations, and domain context.}
    \label{fig: utilization}
\end{figure*}

\subsection{The Prevalence of Charts}
As shown in Fig. \ref{fig: utilization}, charts are a ubiquitous and essential medium for conveying information, as they are highly concise and informative. In the financial sector, charts are common, since visualizations like candlestick charts, line charts, bar charts, and heat maps are crucial in tracking market movements and thus making financial analysis \cite{wang2023finvis}. Furthermore, charts can be expressed in various forms, such as voice descriptions and tactile charts \cite{Moured_2024, 2025-tactile-vega-lite}, to become accessible to the visually impaired. In addition, charts are vital in biomedical image analysis, since they can provide visual summaries, assist in interpreting complex visual representations, and contribute to diagnostic processes \cite{li2023llavamedtraininglargelanguageandvision}. Besides, in scientific research, charts play a pivotal role in illustrating experimental results, new findings, and pipelines, as evidenced by the increasing number of works \cite{lee2016viziometricsanalyzingvisualinformation, reddy2019figurenetdeeplearningmodel, gomezperez2019lookreadenrichlearning, ahmed2023realcqascientificchartquestion, karishma2023aclfigdatasetscientificfigure, shi2024mattersintegrityverificationscientific, wu2024plot2codecomprehensivebenchmarkevaluating, shen2024rethinkingcomprehensivebenchmarkchart, rojas2024enhancingscientificfigurecaptioning, li2025mmscidatasetgraduatelevelmultidiscipline}. On top of that, charts also stand paramount in terms of business management and public policy decision-making, where they are used to present statistics and monitor trends by the companies and governments.

\subsection{MLLM for Chart Understanding}
Before the rise of Multimodal Large Language Models (MLLMs), the downstream tasks of Chart Understanding were mainly restricted to element extraction, chart classification, and naive question answering. These tasks heavily rely on rule-based and heuristic approaches, which often struggle to handle the complexity and diversity of real-world charts. 

For instance, early models usually utilize a fixed-size, limited-volume vocabulary in the output layer to perform question answering tasks. They use LSTM \cite{graves2012long} or Transformer Encoder \cite{vaswani2017attention} to encode textual inputs and a CNN architecture like ResNet \cite{he2016deep} to encode chart inputs. The encoded representations are then fused via concatenation of feature vectors and then passed through a Multi-Layer Perceptron (MLP) to predict the final answer, relying on Softmax Classifier. However, the problem is that these models can only answer questions from a fixed, pre-defined vocabulary that they learned during training. 

Imagine an AI trained only on charts about "apples" and "oranges". If you show it a new chart about "bananas", it would not be able to say the word "banana" because it is never in its original vocabulary. This is the out-of-vocabulary (OOV) problem \cite{kafle2018dvqa}. 

To overcome this obstacle, later models propose Dynamic Encoding to create a vocabulary on the fly for each new chart the models see. Instead of relying on a fixed list, the models read the text directly from the chart images (axis labels and data values) and use those words to form their answers. This approach still struggles to generate the answers accurately because there is an inductive bias that this kind of model assumes that the answers are from the annotated data points on charts, which is not always true. Besides, they can only perform factoid question answering, outputting a brief word rather than a whole sentence. 

However, in the age of MLLMs, these problems no longer exist. To begin with, with a much larger vocabulary and a powerful Language Model, MLLMs are more robust to the OOV problem and more competent with a wider range of downstream tasks. For example, the vocabulary size of GPT is around 100000, far bigger than the vocabulary size of traditional question answering. Due to the Auto-Regressive architecture, MLLMs can generate complete sentences rather than a single word piece, which expands the boundary of downstream tasks. Besides, MLLMs excel at fusing visual and textual modalities within a unified architecture by utilizing Vision Transformer \cite{dosovitskiy2020image} as its Vision Encoder instead of a CNN. Compared to CNNs, Vision Transformer scales better and demonstrates better computational efficiency \cite{radford2021learning}. Furthermore, modern MLLMs demonstrate more powerful reasoning abilities via different prompting strategies, evidenced by Chain of Thought \cite{wei2022chain}, Program of Thought \cite{chen2022program, gao2023palprogramaidedlanguagemodels} and Zero/Few-Shot Learning \cite{brown2020language}, which can help them perform better. \cite{lu2023mathvista, zhang2024mathversedoesmultimodalllm, peng2024multimathbridgingvisualmathematical, jia2024describethenreasonimprovingmultimodalmathematical, sun2025mathglancemultimodallargelanguage, zhu2025mapsadvancingmultimodalreasoning}.

\begin{figure*}
    \centering
    \includegraphics[width=\linewidth]{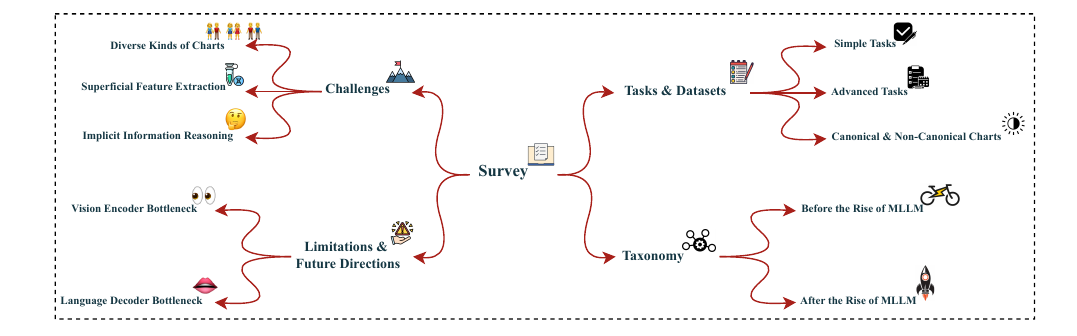}
    \caption{The arrangement of the survey.}
    \label{fig: arrangement}
\end{figure*}

\begin{figure*}
    \centering
    \includegraphics[width=\linewidth]{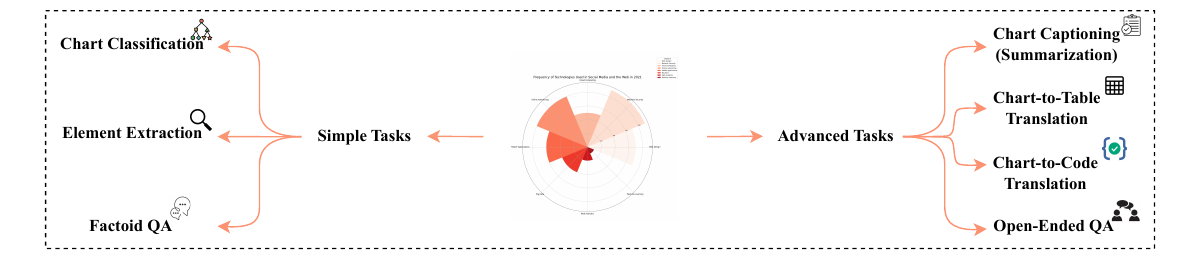}
    \caption{Downstream tasks of chart understanding. These tasks are categorized into simple ones and advanced ones where simple tasks often rely on softmax classifiers or recurrent neural networks (RNNs) to generate outputs, while advanced tasks utilize more powerful auto-regressive language models based on a transformer to generate outputs.}
    \label{fig: tasks}
\end{figure*}

\subsection{Motivation}
Previous surveys, including foundational works \cite{shahira2021towards, shakeel2022comprehensive, hoque2022chartquestionansweringstate, farahani2023automatic, singh2023towards, dhote2023survey, bajic2023review, alshetairy2024transformersutilizationchartunderstanding, huang2024detection} and the notable "From Pixels to Insights: A Survey on Automatic Chart Understanding in the Era of Large Foundation Models" \cite{huang2024pixels}, have provided valuable overviews of Chart Understanding. However, the field is evolving at an unprecedented pace, rendering even recent summaries quickly outdated. While prior works have cataloged the application of large models, they often lack a deep synthesis of the very latest paradigms and benchmarks that have emerged in late 2024 and 2025, neglecting recent achievements in novel datasets and benchmarks \cite{zhang2024mathversedoesmultimodalllm, wei2024mchartqauniversalbenchmarkmultimodal, huang2025evochartbenchmarkselftrainingapproach, masry2025chartqapro} and modeling strategies \cite{moured2024chartformerlargevisionlanguage, xu2024chartmoe, zhou2024chartkgknowledgegraphbasedrepresentationchart, he2024distill, wei2024smalllanguagemodelmeets, zhao2025chartcoder, ji2025socratic, cui2025draw}. For instance, new approaches leveraging reinforcement learning for multi-hop reasoning \cite{jia2025chartreasonercodedrivenmodalitybridging}, dynamic tool selection \cite{huang2025visualtoolagent}, and multi-agent debate for verification \cite{kim2024can} are fundamentally shifting the landscape. Tab. \ref{tab: prior_surveys} summarizes how this survey is positioned against representative prior surveys.

\begin{table*}[t]
    \centering
    \caption{Comparison with representative surveys on chart understanding.}
    \label{tab: prior_surveys}
    \small
    \setlength{\tabcolsep}{4pt}
    \begin{tabular}{@{}>{\raggedright\arraybackslash}p{0.20\textwidth}>{\raggedright\arraybackslash}p{0.40\textwidth}>{\raggedright\arraybackslash}p{0.34\textwidth}@{}}
        \toprule
        Survey & Main focus & Positioning of our survey \\
        \midrule
        Davila et al. \cite{9085944} & Chart mining pipeline, including chart extraction from documents, multi-panel processing, chart classification, data extraction, applications, and datasets. & Extends the scope from chart mining pipelines to MLLM-based multimodal fusion, reasoning, and open-ended analysis. \\
        Hoque et al. \cite{hoque2022chartquestionansweringstate} & Chart question answering, including input/output taxonomy, solution types, datasets, and evaluation techniques. & Covers broader tasks, benchmark taxonomies, open-ended analysis, and reasoning-centric MLLM paradigms. \\
        Farahani et al. \cite{farahani2023automatic} & Automatic chart image understanding, including chart analysis, captioning, and question answering. & Focuses on modern MLLM adaptation, prompting, intermediate representations, tool use, and preference optimization. \\
        Baji{\'c} and Job \cite{bajic2023review}; Dhote et al. \cite{dhote2023survey} & Baji{\'c} and Job review chart-type classification, chart text processing, data extraction, and description generation; Dhote et al. focus on chart classification methods and datasets. & Places these perception and extraction modules as upstream stages within a broader taxonomy of MLLM-based fusion strategies. \\
        Al-Shetairy et al. \cite{alshetairy2024transformersutilizationchartunderstanding} & Transformer-based chart understanding, including E2E frameworks, datasets, evaluation techniques, and pre-training tasks. & Synthesizes later MLLM adaptation, MoE, dynamic tool use, reinforcement learning, and multi-agent reasoning. \\
        Huang et al. \cite{huang2024pixels} & Automatic chart understanding in the era of large foundation models, covering tasks, datasets, metrics, modeling strategies, and tool augmentation. & Adds an information-fusion taxonomy, canonical/non-canonical benchmark analysis, failure-mode diagnosis, and 2025 cognitive enhancement methods. \\
        This survey & MLLM-based chart understanding through a multimodal information-fusion lens. & Systematically connects fusion strategies, chart/task types, limitations, practical trade-offs, and future cognitive enhancement. \\
        \bottomrule
    \end{tabular}
\end{table*}

This survey provides a necessary and timely update by offering three distinct contributions. First, we incorporate a comprehensive review of these cutting-edge 2025 advancements, shedding new light on emergent model architectures, training strategies, and evaluation protocols. Second, we adopt a unique "Information Fusion" lens, analyzing not just what models can do, but how they fuse visual and linguistic information, a perspective crucial for robust system design. Third, we introduce a more granular taxonomy of methodologies and a novel classification of datasets into canonical and non-canonical types, providing a structured roadmap that is absent in prior literature. In light of these circumstances, a new survey is imperative to integrate these recent, transformative advancements. This survey aims to offer a comprehensive overview of Chart Understanding, with an emphasis on how MLLMs are transforming the landscape of this domain.

The structure of the survey is demonstrated in Fig. \ref{fig: arrangement}. To begin with, we first introduce the challenges within the field of Chart Understanding in Sec. \ref{sec: challenges}. In this section, we discuss why charts are hard to interpret. Then we explore the downstream tasks of Chart Understanding, such as chart classification, element extraction, chart captioning, factoid question answering, open-ended question answering, chart-to-table translation, and chart-to-code translation, in Sec. \ref{sec: tasks and datasets}. In this section, we also introduce a bunch of prevailing datasets according to various downstream tasks in order to have a more comprehensive knowledge. We categorize the datasets into canonical and non-canonical ones based on the chart types they encompass. Furthermore, we explore the essential modeling approaches in Sec.  \ref{sec: modeling approaches} before and after the rise of MLLMs, including classification-based models, Generative Models, Pre-Training strategies, Supervised Fine-Tuning strategies, Prompting strategies, augmentation tools, and Reinforcement Learning methods to help the models "see" better. We also introduce methods especially designed for non-canonical charts. Then, in Sec. \ref{sec:limitations}, we meticulously extract the limitations and bottlenecks of the aforementioned approaches and, in Sec. \ref{sec:future_directions}, put forward our own insights into promising future directions. 

\section{Challenges} \label{sec: challenges}
\subsection{Superficial Representation Extraction}
Unlike natural images, which typically capture real-world scenarios with continuous and predictable patterns, charts present a complex and nuanced form of visual representations that intricately interweave graphical elements with textual annotations to convey sophisticated and multifaceted messages. 

Graphical components, such as bars, lines, pie slices, or nodes, encode numerical or categorical data in a highly structured format, while textual elements, including axis labels, legends, titles, and captions, provide essential context, clarify specific data points, or emphasize key insights. This dual modality of visual and textual information demands a comprehensive understanding of both the spatial arrangement of chart elements and the semantic interplay between text and visuals. 

For automated systems like Multimodal Large Language Models (MLLMs), this poses significant challenges, as they must not only process and interpret the visual structure of the chart but also accurately align the textual annotations with the corresponding graphical elements to extract coherent meaning. Failure to effectively fuse these modalities can lead to misinterpretations, as the meaning of a chart often emerges from the precise relationship between its visual and textual components.

\subsection{Implicit Information Reasoning}
Beyond the superficial layer of visual and textual representations, charts often encode intricate and implicit information that is not immediately apparent. While colors, shapes, labels, and legends serve as direct visual cues and textual explanations, the true significance of a chart frequently lies in the patterns, correlations, and trends that emerge when the chart is considered as a whole. 

For instance, a time series line chart may not explicitly display fluctuations or cycles, but it can subtly suggest trends, seasonal patterns, or anomalies that require careful interpretation beyond what is explicitly shown. These implicit insights demand a deeper level of reasoning, as they often rely on understanding the broader context of the data, such as the domain it pertains to or the intended message of the chart.

For MLLMs, extracting and interpreting this implicit information is particularly challenging, as it requires not only recognizing the visible elements but also fusing them to uncover underlying relationships and meanings that are not directly encoded in the chart’s visual or textual components.

\subsection{Diverse Kinds of Chart}
Charts encompass a wide variety of types, including bar charts, line charts, pie charts, scatter plots, 3D plots, flowcharts, block diagrams, box plots, tree diagrams, heatmaps, parallel coordinate plots, rose charts, radar charts, bubble charts, multi-axis charts, multi-chart dashboards, funnel charts, candlestick charts, choropleth charts, contour plots, sankey charts, and more, each defined by distinct external representations and internal semantic conventions that present substantial challenges for MLLMs to parse and comprehend.

Each chart type employs unique visual structures that are tailored to represent specific kinds of data, whether categorical, numerical, or relational. These visual forms are further complicated by varying semantic conventions, such as axis labels, legends, color coding, or annotations, which convey meaning specific to the chart type and its intended use. 

This diversity in both visual form and semantic interpretation demands that MLLMs not only recognize and differentiate between graphical patterns but also contextualize them within the appropriate semantic framework, adapting their fusion and reasoning strategies to the conventions of each chart type. The ability to generalize across such varied representations while maintaining accuracy in interpretation remains a significant hurdle for automated systems aiming to fully understand and analyze charts.

\renewcommand{\arraystretch}{2}
\begin{table*}
    \centering
    \footnotesize
    \begin{tabular}{c c c c c c}
         \hline
         Dataset & Task & Quantity & Image Source & Text Source & Non-Canonical \\
         \hline
         FigureQA \cite{kahou2017figureqa} & FQA & 1.6M & \makecell{Matplotlib} & Templates & \ding{55} \\
         DVQA \cite{kafle2018dvqa} & FQA & 3.4M & \makecell{Matplotlib} & Templates & \ding{55} \\
         PlotQA \cite{methani2020plotqa} & FQA & 28.9M & \makecell{Matplotlib, \\ World Bank} & Templates & \ding{55} \\
         LEAF-QA \cite{chaudhry2020leaf} & FQA & 2M & \makecell{Matplotlib, \\ World Bank} & Templates & \ding{55} \\
         Chart2Text \cite{obeid2020chart} & CAP & 8K & Statista & Statista & \ding{55} \\
         STL-CQA \cite{singh2020stl} & FQA & Undisclosed & LEAF-QA \cite{chaudhry2020leaf} & \makecell{LEAF-QA \cite{chaudhry2020leaf}, \\ Templates} & \ding{55} \\
         DocVQA \cite{mathew2021docvqa} & FQA & 50K & UCSF & Human Anno. & \ding{51} \\
         AutoChart \cite{zhu2021autochartdatasetcharttotextgeneration} & CAP & 23K & Matplotlib & Templates & \ding{55} \\
         InfographicVQA \cite{mathew2021infographicvqa} & FQA & 30K & Internet & Human Anno. & \ding{51} \\
         SCICAP \cite{hsu2021scicap} & CAP & 2M & arXiv & arXiv & \ding{51} \\
         IconQA \cite{lu2021iconqa} & FQA/OQA & 107K &  IXL Math Learning & Human anno. & \ding{51} \\
         Chart-to-Text \cite{kantharaj2022chart} & CAP & 44K & \makecell{Statista, Pew} & \makecell{Statista, Pew} & \ding{55} \\
         ChartQA \cite{masry2022chartqa} & FQA & 32K & \makecell{Statista, Pew, \\ OWID, OECD} & \makecell{Human Anno., \\ Fine-Tuned T5} & \ding{55} \\
         OpenCQA \cite{kantharaj2022opencqa} & OQA & 7.7K & Pew & Human Anno. & \ding{55} \\
         MapQA \cite{chang2022mapqadatasetquestionanswering} & FQA & 800K & \makecell{KFF, GeoPandas, \\ Plotly} & Templates & \ding{51} \\
         SlideVQA \cite{tanaka2023slidevqa} & FQA/OQA & 14K & SlideShare & Human Anno. & \ding{51} \\
         ChartSumm \cite{rahman2023chartsumm} & CAP & 84.3K & \makecell{Knoema, Statista} & \makecell{Knoema, Statista} & \ding{55} \\
         VisText \cite{tang2023vistext} & CAP & 12.4K & Statista, Python & Human Anno. & \ding{55} \\
         SciCap+ \cite{yang2023scicapknowledgeaugmenteddataset} & CAP & 414K & SciCap \cite{hsu2021scicap} & SciCap \cite{hsu2021scicap} & \ding{51} \\
         SciGraphQA \cite{li2023scigraphqa} & FQA/OQA & 296K & SciCap+ \cite{yang2023scicapknowledgeaugmenteddataset} & PaLM 2 \cite{anil2023palm2technicalreport}, GPT-4 \cite{achiam2023gpt} & \ding{51} \\
         FlowchartQA \cite{tannert2023flowchartqa} & FQA & 6M & Graphviz & Templates & \ding{51} \\
         MMC-Instruction \cite{liu2023mmc} & \makecell{FQA/OQA/CAP \\ C2T/C2C} & 600K & \makecell{arXiv, Statista, \\ PlotQA \cite{methani2020plotqa}, VisText \cite{tang2023vistext}} & \makecell{arXiv, Statista, GPT-4 \cite{achiam2023gpt}, \\ PlotQA \cite{methani2020plotqa}, VisText \cite{tang2023vistext}} & \ding{55} \\
         \hline
    \end{tabular}
    \caption{An overview of the most prevalent chart-specific datasets. FQA refers to factoid question answering, OQA refers to open-ended question answering, CAP refers to captioning, C2T refers to chart-to-table translation, and C2C refers to chart-to-code translation.}
    \label{tab: overview of prevalent datasets1}
\end{table*}

\renewcommand{\arraystretch}{2}
\begin{table*}
    \centering
    \footnotesize
    \begin{tabular}{c c c c c c}
         \hline
         Dataset & Task & Quantity & Image Source & Text Source & Non-Canonical \\
         \hline
         M-Paper \cite{hu2024mplug} & OQA/CAP & 300K & arXiv & arXiv  & \ding{51} \\
         ChartBench \cite{xu2023chartbench} & FQA & 600K & Kaggle, Matplotlib & GPT-3.5 & \ding{51} \\
         ChartSFT \cite{meng2024chartassisstant} & \makecell{FQA/OQA \\ CAP/C2T} & 39M & \makecell{arXiv, ChartQA \cite{masry2022chartqa}, etc.} & \makecell{arXiv, ChartQA \cite{masry2022chartqa}, etc.} & \ding{55} \\
         ChartX \cite{xia2024chartx} & \makecell{FQA/CAP/C2T/C2C} & 48K & \makecell{GPT-4 \cite{achiam2023gpt}, Human Superv.} & \makecell{GPT-4 \cite{achiam2023gpt}, \\ Human Superv.} & \ding{51} \\
         ChartInstruct \cite{masry2024chartinstruct} & FQA/CAP/C2T/C2C & 191K & \makecell{UniChart \cite{masry2023unichart}} & \makecell{GPT-3.5 Turbo, \\ GPT-4} & \ding{55} \\
         ChartMimic \cite{yang2025chartmimicevaluatinglmmscrossmodal} & C2C & 4.8K & \makecell{arXiv, Stack Overflow, \\ Matplotlib Gallery, etc.} & Human Anno. & \ding{51} \\
         Plot2Code \cite{wu2024plot2codecomprehensivebenchmarkevaluating} & C2C & 368 & \makecell{Matplotlib Gallery, \\ Plotly} & \makecell{Matplotlib Gallery, \\ Plotly, GPT-4 \cite{achiam2023gpt}} & \ding{55} \\
         CharXiv \cite{wang2024charxiv} & FQA & 11K & arXiv & \makecell{GPT-4V, \\ Human Anno.} & \ding{51} \\
         ArXivCap \cite{li2024multimodal} & CAP & 3.9M & \makecell{arXiv, \\ Semantic Scholar} & \makecell{arXiv, \\ Semantic Scholar} & \ding{51} \\
         ArXivQA \cite{li2024multimodalarxivdatasetimproving} & FQA/OQA & 100K & \makecell{arXiv, \\ Semantic Scholar} & \makecell{arXiv, GPT-4V, \\ Semantic Scholar} & \ding{51} \\
         MMSCI \cite{li2025mmscidatasetgraduatelevelmultidiscipline}& CAP & 742K & Nature Commu. & Nature Commu. & \ding{51} \\
         FlowVQA \cite{singh2024flowvqamappingmultimodallogic} & FQA & 22K & \makecell{WikiHow, FloCo, \\ Instructables DIY} & GPT-4, Templates & \ding{51} \\
         NovaChart \cite{hu2024novachart} & \makecell{FQA/OQA/CAP \\ C2T/C2C} & 856K & Python & GPT-4 & \ding{51} \\
         ChartBank \cite{yang2024askchartuniversalchartunderstanding} & FQA/OQA/CAP/C2T & 7.5M & \makecell{ChartQA \cite{masry2022chartqa}, \\ UniChart \cite{masry2023unichart}, etc.} & ChatGPT \cite{ouyang2022traininglanguagemodelsfollow} & \ding{55} \\
         Chart2Code \cite{zhao2025chartcoder} & C2C & 160K & GPT-4o & GPT-4o & \ding{55} \\
         MultiChartQA \cite{zhu2024multichartqa} & FQA/OQA & 2K & \makecell{Data Commons, \\ GALLUP, \\ USAFacts, etc.} & Human Anno. & \ding{51} \\
         RefChartQA \cite{vogel2025refchartqagroundingvisualanswer} & FQA & 73K & ChartQA \cite{masry2022chartqa} & GPT-4o & \ding{55} \\
         ChartQAPro \cite{masry2025chartqapro} & FQA/OQA & 1.9K & \makecell{OWID, PPIC, \\ Tableau, Pew} & \makecell{Human Anno., \\ GPT-4o} & \ding{51} \\
         \hline
    \end{tabular}
    \caption{An overview of the most prevalent chart-specific datasets. FQA refers to factoid question answering, OQA refers to open-ended question answering, CAP refers to captioning, C2T refers to chart-to-table translation, and C2C refers to chart-to-code translation.}
    \label{tab: overview of prevalent datasets2}
\end{table*}

\section{Tasks and Datasets} \label{sec: tasks and datasets}
\subsection{Simple Tasks}
\subsubsection{Chart Classification}
Fig. \ref{fig: tasks} showcases the downstream tasks of Chart Understanding. Chart classification \cite{8489315, 9085944, 10.1145/3469096.3474931, dhote2023survey, bajic2023review}, serving as the cornerstone of Chart Understanding, is the pivotal task of assigning a given chart to one of several predefined categories. The complexity arises from the vast diversity of chart types, each with its unique visual grammar and purpose. Researchers leverage advanced Computer Vision techniques, often employing Deep Learning architectures such as Convolutional Neural Networks (CNNs), to learn intricate features from chart images.

\subsubsection{Element Extraction}
Element extraction \cite{gao2012view, battle2017beagleautomatedextractioninterpretation, Cliche_2017, liu2019data, luo2021chartocr, Yan_2023, Mustafa_2023, xue2023chartdetrmultishapedetectionnetwork, huang2024detection}, often referred to as element detection, delves into the meticulous process of identifying and recognizing the constituent visual components within a chart. The techniques employed often involve object detection algorithms, such as You Only Look Once (YOLO) \cite{redmon2016lookonceunifiedrealtime} or Faster R-CNN \cite{ren2016fasterrcnnrealtimeobject}, trained on extensive datasets of annotated charts. The primary challenge lies in handling the variability in element appearances, sizes, and orientations. Beyond mere detection, recognition involves interpreting the meaning of these elements.

\subsubsection{Factoid QA}
Factoid question answering on charts represents a critical step towards enabling intelligent interaction with visual data. Unlike open-ended queries, factoid questions are designed to elicit precise, objective single-word answers directly derivable from the chart's content. 

Consider a bar chart displaying sales figures for different regions: A factoid question might be "What were the sales in the North region?" or "Which region had the highest sales?". The system's ability to answer these questions relies on its capacity to accurately extract numerical values, identify categorical labels, and perform simple comparative or aggregative operations on the retrieved data. This often involves a pipeline that first extracts the underlying data from the chart and then employs Natural Language Processing (NLP) techniques to understand the question and query the extracted data. 

\subsection{Advanced Tasks}
\subsubsection{Chart Captioning}
Chart captioning, also known as chart summarization, elevates Chart Understanding to the realm of natural language generation. This advanced task aims to automatically generate a concise, and informative textual description that encapsulates the key insights presented in a given chart. It goes far beyond simply listing the chart's elements; it requires the system to comprehend the data's narrative. This necessitates a deep understanding of the data's patterns, anomalies, and relationships. Techniques often involve a combination of Computer Vision for chart analysis, data extraction, and natural language generation models.

\subsubsection{Chart-to-Table Translation}
Chart-to-table translation is a transformative task that bridges the gap between visual data representation and structured, machine-readable data. The core objective is to convert the graphical information embedded within a chart into a well-organized tabular format, where rows represent records and columns represent attributes. This process involves not only element extraction but also a crucial step of data inference, where the system must accurately interpret scales, units, and relationships between visual elements to reconstruct the underlying numerical data. 

\subsubsection{Chart-to-Code Translation}
Chart-to-code translation is an ambitious and highly practical task that aims to reverse-engineer the process of chart creation. The goal is to generate executable code (e.g., Python code using libraries like Matplotlib or Seaborn, XML, SVG, or other visualization-specific languages) that, when executed, can precisely reproduce the input chart. This task demands a deep understanding of the chart's visual grammar, underlying data, and the specific parameters that govern its rendering. For a given bar chart, the system would need to identify the data values, the bar colors, the axis labels, the title, the legend, and then translate all these visual attributes into corresponding code commands.

\subsubsection{Open-Ended QA}
Open-ended question answering on charts pushes the boundaries of chart comprehension beyond simple factual retrieval. Unlike factoid questions, which seek a specific, factual answer (often a single word or number), open-ended queries invite more descriptive, multi-sentence responses. The system must not only extract data but also infer meaning, identify trends, and synthesize information to provide nuanced answers.

The system needs to understand not just what the data says, but why it says it, and what it implies. This often involves integrating Chart Understanding with broader knowledge bases and leveraging advanced natural language understanding and generation techniques. The primary challenges lie in handling the inherent ambiguity of open-ended questions, generating relevant and insightful responses.

\subsection{Canonical and Non-Canonical Charts}
\begin{figure*}
    \centering
    \includegraphics[width=\linewidth]{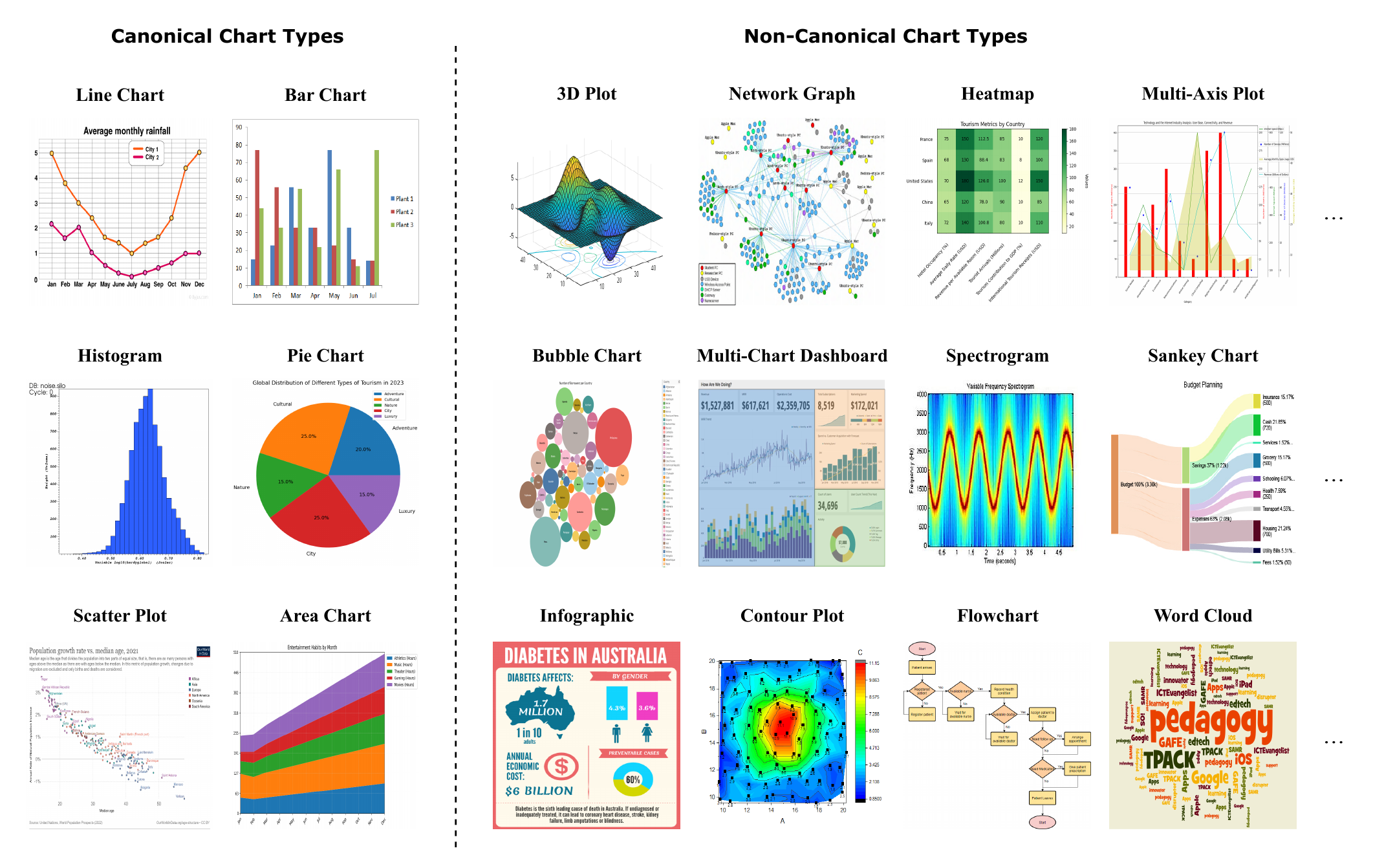}
    \caption{Examples of canonical and non-canonical charts. Canonical charts include line charts, bar charts, histograms, and pie charts, scatter plots, and area charts. Non-canonical charts encompass a wide range of chart types, including and not limited to the charts illustrated in the figure above.}
    \label{fig: canonical}
\end{figure*}

As shown in Fig. \ref{fig: canonical}, Chart Understanding has predominantly focused on a limited subset of chart types, namely bar charts, line charts, pie charts, scatter plots, area charts, and histograms. These types, often referred to as statistical charts, constitute the backbone of many benchmark datasets such as ChartQA \cite{masry2022chartqa}, PlotQA \cite{methani2020plotqa}, and DVQA \cite{kafle2018dvqa}. While these visualizations are indeed common in articles, dashboards, and reports, they represent only a small fraction of the wide spectrum of visual representations in real-world contexts. This narrow focus creates a significant gap between the capabilities of current Chart Understanding models and the rich visual diversity found in the wild.

In real-world scenarios, data visualizations go far beyond the statistical archetypes mentioned above. There exists a broad and diverse range of chart types that remain underexplored or completely absent in many existing datasets and benchmarks. These include, but are not limited to box plots, flowcharts, tree diagrams, network graphs, 3D plots, heatmaps, parallel coordinate plots, rose charts, radar charts, bubble charts, multi-axis plots, multi-chart dashboards, funnel charts, candlestick charts, choropleth maps, contour plots, sankey charts, word clouds, knowledge graphs, and domain-specific formats such as spectrograms used in audio analysis or phylogenetic trees in biology. 

A particularly noteworthy category is infographics, which frequently integrate multiple chart types, textual annotations, and illustrative icons within a single visual narrative. Understanding these complex chart types poses substantial challenges. They often require richer multimodal reasoning, including the integration of visual structure, textual semantics, and domain-specific knowledge. Additionally, many of these charts feature non-standard layouts, dense labeling, and semantic ambiguity, all of which degrade the performance of models that are primarily trained on clean, uniform, and narrowly-scoped statistical charts. For example, a single infographic may embed several different types of charts, such as a pie chart, line graph, and choropleth map, each representing a distinct aspect of a complex phenomenon (e.g., climate change, public health crises, economic inequality). Proper interpretation requires cross-chart reasoning and the ability to synthesize information distributed across different visual elements. Moreover, infographics often use artistic fonts, making text recognition challenging even for state-of-the-art systems. Besides, They may also incorporate graphical icons, illustrations, and stylized decorations to enhance visual appeal or storytelling, further increasing the cognitive load required for accurate interpretation. These elements introduce non-linguistic visual semantics that must be resolved in tandem with textual and numerical cues. 

Furthermore, some chart types rely heavily on spatial relationships. For example, node-link diagrams in network graphs, hierarchical depth in tree diagrams, or positional encoding in 3D surface plots. In these cases, a model must not only parse individual elements but also reason over their spatial arrangements and inter-element dependencies.

In summary, the current landscape of Chart Understanding is constrained by an overreliance on a narrow class of statistical chart types, failing to address the full visual and cognitive complexity of charts used in real-world communication. Bridging this gap will require the development of new benchmarks, richer multimodal pretraining, and more robust model architectures that can handle layout variability, visual diversity, and domain adaptation at scale.

Tab. \ref{tab: overview of prevalent datasets1} and Tab. \ref{tab: overview of prevalent datasets2} demonstrate the most prevalent datasets widely used in Chart Understanding. Recently, an increasing number of works have emerged to extend Chart Understanding beyond canonical statistical charts toward a broader spectrum of non-canonical chart types. These efforts include the construction of diverse datasets.

One major direction involves scientific visualizations that appear in academic publications or experimental reports. These charts typically contain specialized notations, and dense annotations. To address this domain, a number of datasets and benchmarks have been proposed. For instance, SCICAP \cite{hsu2021scicap, hsu2021scicapgeneratingcaptionsscientific} focuses on generating captions for scientific figures, while RealCQA \cite{ahmed2023realcqascientificchartquestion} and SciGraphQA \cite{li2023scigraphqa} provide question answering settings for interpreting scientific charts. ChartQAPro \cite{masry2025chartqapro} and SCI-CQA \cite{shen2024rethinkingcomprehensivebenchmarkchart} further expand the benchmark landscape to support broader real-world chart QA scenarios.

Another prominent line of work deals with infographic-style visualizations. These infographics go beyond the isolated chart paradigm and require models to perform layout reasoning, visual-textual grounding, and cross-chart synthesis. To address this complexity, several datasets have been introduced. DocVQA \cite{mathew2021docvqa} and InfographicVQA \cite{mathew2021infographicvqa} were among the first to explore QA tasks on infographic documents. More recently, SlideVQA \cite{tanaka2023slidevqa, tanaka2023slidevqadatasetdocumentvisual} focuses on Chart Understanding in slide-based contexts, and MultiChartQA \cite{zhu2024multichartqa} explicitly addresses the challenge of reasoning over multiple co-located charts. ChartGalaxy \cite{li2025chartgalaxydatasetinfographicchart} further enriches the field with a large-scale dataset of infographic-style charts. 

Map-related visualizations also represent an underexplored category. These include choropleth maps and geospatial encodings where meaning is embedded in color gradients, shapes, and location-sensitive structures. The MapQA dataset \cite{chang2022mapqadatasetquestionanswering} introduces a benchmark for question answering tasks over maps. In parallel, a growing body of research has turned attention to flowchart-based visualizations, which often embody procedural logic, and hierarchical dependencies. These charts present unique challenges due to their graph-based structures, and directional flows. FlowLearn \cite{pan2024flowlearnevaluatinglargevisionlanguage} and FlowVQA \cite{singh2024flowvqamappingmultimodallogic} introduce benchmarks and evaluation frameworks for understanding logical reasoning in flowcharts.


Collectively, these works highlight a growing recognition of the limitations in traditional Chart Understanding paradigms and reflect a broader shift toward tackling the structural diversity, and reasoning complexity inherent in non-canonical visualizations. They also lay the groundwork for future research in Chart Understanding that is more generalizable, and applicable across diverse real-world scenarios.

\subsection{Benchmark Bias and Evaluation Reliability}
The increasingly diverse benchmark landscape also makes cross-dataset comparison less straightforward. As summarized in Tab. \ref{tab:benchmark_bias}, datasets differ not only in scale and chart type, but also in visual source, annotation procedure, question form, and evaluation metric. Synthetic benchmarks provide controllable supervision and large-scale training signals, but they may overrepresent clean layouts and templated language. In contrast, realistic, scientific, and non-canonical benchmarks better reflect practical use cases, yet they often entangle visual grounding, OCR, domain knowledge, and multi-step reasoning within a single answer score.

\begin{table*}[t]
    \centering
    \footnotesize
    \renewcommand{\arraystretch}{1.25}
    \setlength{\tabcolsep}{3pt}
    \begin{tabular}{@{}>{\raggedright\arraybackslash}p{0.15\linewidth}>{\raggedright\arraybackslash}p{0.22\linewidth}>{\raggedright\arraybackslash}p{0.18\linewidth}>{\raggedright\arraybackslash}p{0.18\linewidth}>{\raggedright\arraybackslash}p{0.21\linewidth}@{}}
        \hline
        Benchmark type & Representative datasets & Evaluation focus & Typical metric & Main bias or limitation \\
        \hline
        Synthetic canonical QA & FigureQA \cite{kahou2017figureqa}, DVQA \cite{kafle2018dvqa}, PlotQA \cite{methani2020plotqa}, LEAF-QA \cite{chaudhry2020leaf}  & Value lookup, comparison, and arithmetic over regular plots & Exact or relaxed answer accuracy & Large scale and clean labels, but template-driven questions and regular layouts can encourage pattern matching. \\
        Real-world and broad statistical QA & ChartQA \cite{masry2022chartqa}, ChartBench \cite{xu2023chartbench}, RefChartQA \cite{vogel2025refchartqagroundingvisualanswer}, ChartQAPro \cite{masry2025chartqapro} & Human-like analytical questions, broad chart QA, and visual answer grounding & Answer accuracy, Acc+, and grounding metrics & More realistic sources and broader coverage, but answer-level scores may not reveal whether errors come from perception, reasoning, grounding, or language generation. \\
        Non-canonical QA & CharXiv \cite{wang2024charxiv}, ArXivQA \cite{li2024multimodalarxivdatasetimproving}, MultiChartQA \cite{zhu2024multichartqa}, FlowchartQA \cite{tannert2023flowchartqa} & Scientific figures, multi-chart layouts, and flowchart logic & Accuracy or semantic grading, depending on the task & Closer to real use cases, but domain knowledge, OCR, topology recovery, and cross-region synthesis are difficult to separate. \\
        Captioning and derendering & ChartSumm \cite{rahman2023chartsumm}, ChartX \cite{xia2024chartx}, DEPLOT \cite{liu2022deplot}, Plot2Code \cite{wu2024plot2codecomprehensivebenchmarkevaluating} & Textual summaries, table recovery, and chart-to-code generation & Text-overlap or semantic metrics; RMS; code pass rate and rendered-image rating & Text metrics may miss factual faithfulness, while structural or rendering metrics can be sensitive to representation choices. \\
        \hline
    \end{tabular}
    \caption{A high-level comparison of benchmark design choices, evaluation metrics, and potential biases in chart understanding.}
    \label{tab:benchmark_bias}
\end{table*}

These differences suggest that benchmark scores should be interpreted as task-specific indicators rather than direct evidence of general chart intelligence. Exact or relaxed accuracy is appropriate for factoid QA with short numeric or categorical answers, but it cannot fully expose whether a model relied on OCR shortcuts, visual grounding, or genuine reasoning. Text-overlap metrics used in captioning and summarization are efficient, but they may reward lexical similarity without verifying whether the generated statement is visually faithful. Structure-oriented metrics such as RMS for table recovery, or code execution and rendered-image comparison for chart-to-code generation, provide stronger evidence for representational fidelity, but they still do not fully capture open-ended semantic interpretation. Future evaluations should therefore report results by chart type, source domain, answer format, and reasoning requirement, and should complement aggregate accuracy with perturbation, label-masking, or counterfactual tests that better distinguish visual reasoning from pattern matching.

\section{Taxonomy} \label{sec: modeling approaches}
\begin{figure*}
    \centering
    \includegraphics[width=\linewidth]{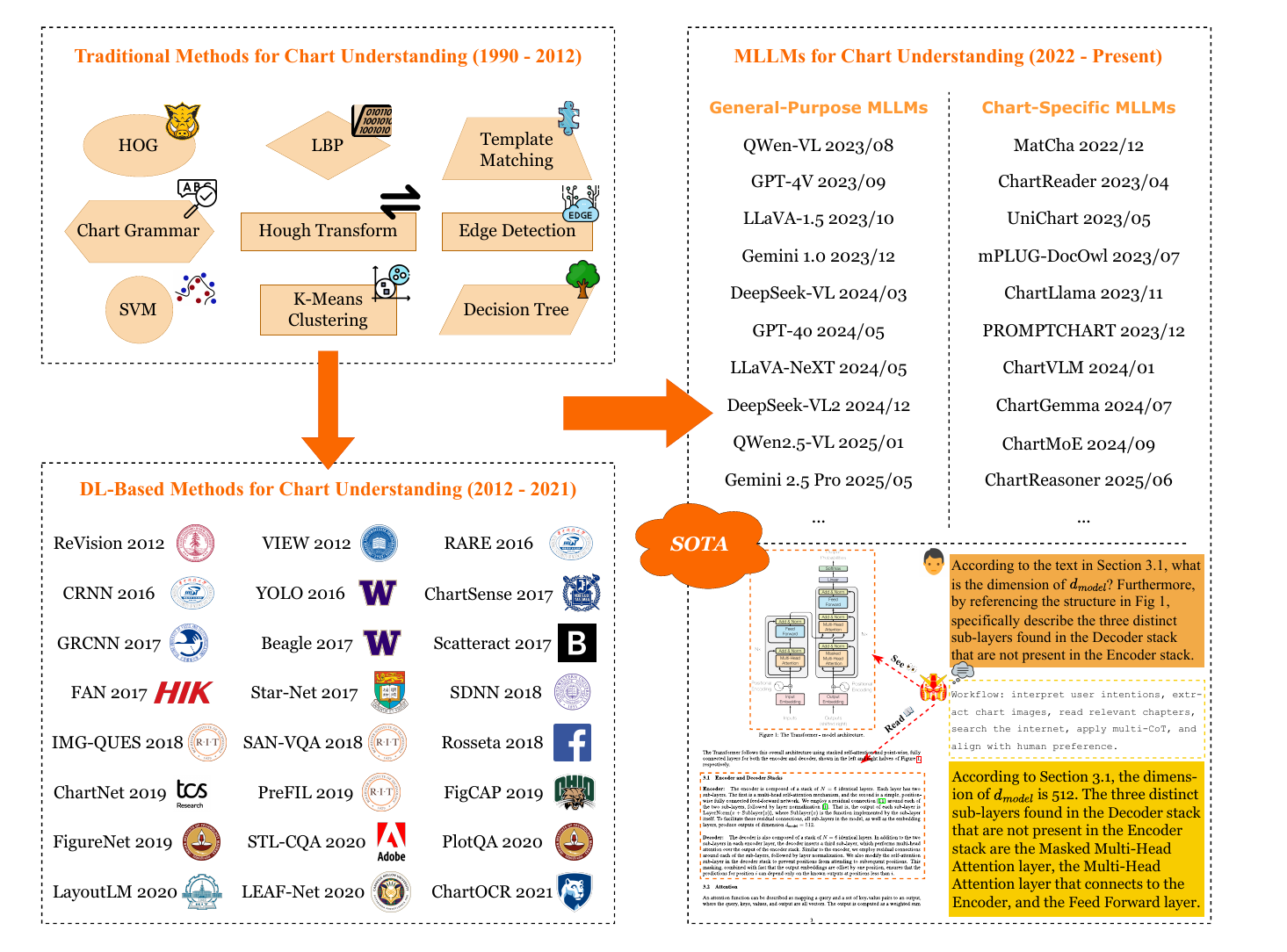}
    \caption{A high-level roadmap of chart understanding methods, showing the transition from traditional CV/ML and deep learning pipelines to MLLM-based fusion paradigms and a prospective scientific reading scenario.}
    \label{fig: tax}
\end{figure*}

Fig. \ref{fig: tax} summarizes the methodological evolution of Chart Understanding across several stages. Early systems relied on handcrafted visual features and rule-based recognition, including Histogram of Oriented Gradients (HOG) \cite{article}, Local Binary Patterns (LBP) \cite{10.5555/2102160.2102213}, template matching \cite{953947}, chart grammar \cite{520809, 10.1145/108360.108361}, Hough transform \cite{10.1145/1284420.1284427}, edge detection \cite{inproceedings}, and decision trees \cite{inproceedings1}. With the rise of deep learning, the field shifted from handcrafted features toward learned visual representations, where CNNs \cite{krizhevsky2012imagenet}, R-CNN \cite{girshick2014richfeaturehierarchiesaccurate}, and Faster R-CNN \cite{ren2016fasterrcnnrealtimeobject} were used to extract visual features and localize chart elements. This stage led to chart-oriented systems such as ReVision \cite{10.1145/2047196.2047247}, VIEW \cite{gao2012view}, RARE \cite{shi2016robust}, CRNN \cite{shi2016end}, and ChartSense \cite{10.1145/3025453.3025957}, as well as early chart QA models such as SAN-VQA and IMG-QUES \cite{kafle2018dvqa}. Later document and vision-language models, such as LayoutLM \cite{xu2020layoutlm}, began to integrate visual layout and textual content more explicitly, preparing the transition toward MLLM-based chart understanding.

The rightmost part of Fig. \ref{fig: tax} sketches a prospective application. In such a scenario, MLLMs could support scientific reading by linking chart interpretation with surrounding document context, citation evidence, and user queries, thereby assisting literature review and knowledge synthesis.

\subsection{Before the Rise of MLLM}
Before the emergence of Multimodal Large Language Models (MLLMs), the domain of Chart Understanding was predominantly navigated by methodologies rooted in Computer Vision (CV). During this era, the immediate objectives and downstream applications of Chart Understanding systems were largely confined to these well-defined tasks: the classification of chart types, the precise extraction of individual chart elements, and factoid question answering that aimed to retrieve specific, explicit pieces of information from the chart. This section focuses on DL-based methodologies.

The computational backbone for visual encoding in these models typically leveraged the power of Convolutional Neural Networks (CNNs), with architectures like ResNet \cite{he2016deep} serving as feature extractors to capture visual representations of the chart images. For the generation of textual outputs, these systems commonly employed Recurrent Neural Networks (RNNs), particularly variants like Long Short-Term Memory (LSTM) networks \cite{graves2012long}, which excelled at processing sequential data to produce coherent captions or concise answers \cite{shi2016end, shi2016robust, liu2016star, wang2017gated}. This modular design, while demonstrably effective for addressing specific, and constrained tasks, often struggles with the holistic and nuanced interpretation that modern MLLMs now aspire to achieve.

\begin{figure*}
    \centering
    \includegraphics[width=\linewidth]{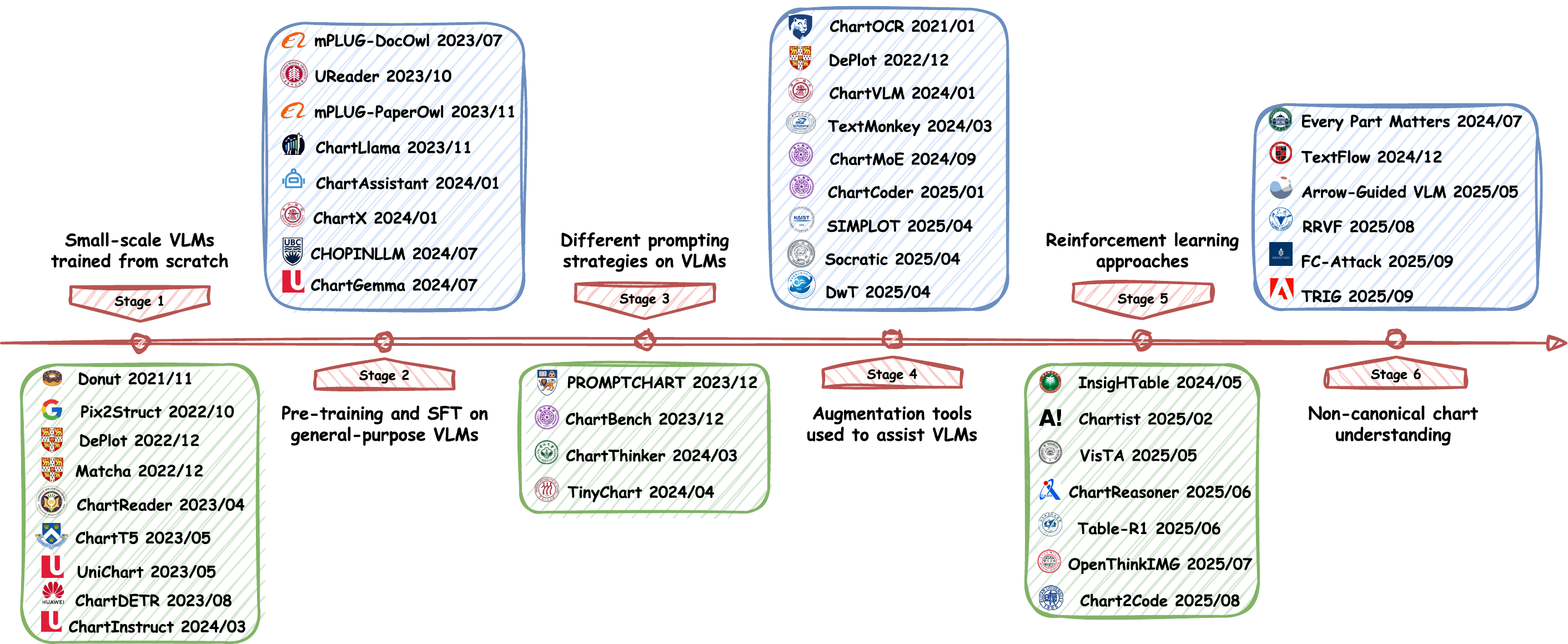}
    \caption{A timeline of advancements in MLLM for Chart Understanding.}
    \label{fig: timeline}
\end{figure*}

\subsubsection{Chart Derendering} Chart derendering aims to reverse-engineer the visualization pipeline, reconstructing the underlying data from a raster image. This process is typically decomposed into two modular stages: classification, and element detection.

\textbf{CNNs.} Before the dominance of object detection methods, Convolutional Neural Networks (CNNs) served as the primary feature extractors for high-level chart classification. Early frameworks, such as ReVision \cite{10.1145/2047196.2047247} and implementations within the FigureQA \cite{kahou2017figureqa} baseline, utilized standard backbones (e.g., VGG-16, ResNet-50) to categorize chart images by type. These classifications were requisite precursors, determining which downstream rule-based heuristics to apply. While effective for classification, global CNN features lacked the spatial granularity required to disentangle dense data points, necessitating the shift toward region-based architectures.

\textbf{Object Detection.} To address the spatial ambiguity of global features, researchers adopted object detection paradigms to localize atomic chart elements. Scatteract \cite{Cliche_2017} pioneered this by fine-tuning R-CNN architectures to extract key components such as axes and data series. This granular approach was extended by Chart-Text \cite{balaji2018charttextfullyautomatedchart}, which employed specialized detectors for titles, labels, and legends. Notably, the field later evolved from standard bounding-box detection (e.g., Faster R-CNN) to keypoint estimation techniques. For instance, ChartOCR \cite{luo2021chartocr} leveraged CornerNet \cite{law2019cornernetdetectingobjectspaired} to detect the precise corners of bars and line vertices, significantly improving data recovery accuracy by treating chart elements as geometric keypoints rather than generic objects.

\textbf{OCR.} Optical Character Recognition (OCR) in Chart Understanding transcends simple text transcription. Instead, it requires rigorous visual-semantic alignment. Early systems relied on off-the-shelf engines like Tesseract, which often failed to effectively map text to its corresponding visual marker (e.g., associating a legend label with a specific color). Advanced methodologies, such as LEAF-Net \cite{chaudhry2020leaf} and finding within ChartSense \cite{10.1145/3025453.3025957}, integrated OCR directly into the deep learning pipeline. These approaches utilized end-to-end differentiable networks to simultaneously localize text and project it into the semantic space of the chart, thereby resolving ambiguities in axis-tick alignment and legend association.

\subsubsection{Factoid Question-Answering} Unlike chart derendering, factoid question answering (FQA) focuses on direct query response. Pre-MLLM methodologies largely treated this as a Visual Question Answering (VQA) task, relying on specialized inductive biases to reason over chart images.

\textbf{Fixed Vocabulary (Classification-Based VQA).} Prior to generative language models, chart QA was predominantly framed as a multi-class classification problem. Models trained on datasets like DVQA \cite{kafle2018dvqa} and FigureQA \cite{kahou2017figureqa} utilized a "closed-set" assumption, where the output was a softmax probability distribution over a pre-defined dictionary of frequent answers (e.g., colors, "yes/no", specific integers). While computationally efficient, this paradigm suffered from rigid inflexibility. Models like PReFIL \cite{kafle2020answeringquestionsdatavisualizations} could accurately classify structural queries but failed when questions required extracting out-of-vocabulary (OOV) tokens, such as specific data values or axis labels explicitly rendered in the image.

\textbf{CNN-LSTM-Like Architectures.} The standard architectural baseline for this era employed a "Late Fusion" mechanism. These models utilized independent encoders: a CNN (e.g., ResNet) to encode visual features and a Recurrent Neural Network (e.g., RNN/LSTM) to embed the textual query. The resulting vectors were concatenated or multiplied element-wise to predict the answer. To enhance reasoning capabilities, subsequent works integrated spatial attention mechanisms, such as Stacked Attention Networks (SAN) \cite{yang2016stackedattentionnetworksimage}, enabling the model to dynamically focus on relevant chart regions conditioned on the question.

\textbf{Structure-Aware and Hybrid Models.} Recognizing the limitations of simple fusion, later architectures attempted to model the relational structure of charts. Relation Networks (RNs) were introduced to explicitly compute pairwise dependencies between detected objects, resolving comparative queries (e.g., "Is bar A higher than bar B?") \cite{santoro2017simpleneuralnetworkmodule,kahou2017figureqa}. This line of work naturally connects to graphical inference models, where chart elements and question tokens are treated as structured entities rather than only as global image features. For example, STL-CQA explicitly localizes chart components and encodes their structure for chart question answering \cite{singh2020stl}, VizExtract extracts relations from statistical visualizations \cite{decatur2021vizextractautomaticrelationextraction}, and ChartKG represents visual marks, textual entities, and semantic links as a knowledge graph \cite{zhou2024chartkgknowledgegraphbasedrepresentationchart}. More recent graph-based CQA further converts chart-question pairs into heterogeneous graphs and performs graph neural network-based inference \cite{10.1016/j.eswa.2025.126508}. These neuro-symbolic and graphical inference models provide a non-LLM alternative in which neural perception modules handle element detection and OCR, while symbolic or graph-based modules perform relation recovery, graph traversal, or structured reasoning. Furthermore, to overcome the fixed vocabulary constraint, hybrid systems like PlotQA \cite{methani2020plotqa} and ChartNet \cite{Sharma_2019} introduced pointer-generator networks and dynamic OCR-copy mechanisms. These innovations allowed models to "copy" text directly from the image into the output sequence, bridging the gap between classification-based VQA and modern open-ended generation.


\subsection{After the Rise of MLLM}
The aforementioned approaches mainly handle simple tasks, such as chart classification, element extraction, and naive question answering. With the rapid development of MLLMs, Chart Understanding has expanded from recognition-oriented tasks toward generation and reasoning tasks, including chart captioning, chart-to-table generation, chart-to-code generation, and open-ended QA. 


Figure \ref{fig: timeline} demonstrates the timeline of advancements of MLLM for Chart Understanding. To reduce overlap among related methods, this section organizes recent progress by the role each paradigm plays in the fusion pipeline. Component-assembled VLMs establish the basic vision-language architecture for charts. Adapting general-purpose MLLMs improves alignment and reasoning capacity through chart-specific tuning or distillation. Prompting and inference-time reasoning then exploit the in-context learning ability of these models without updating parameters. Intermediate representations address information loss by converting charts into tables, code, SVG/XML, or dynamically selected representations. Reinforcement learning and preference optimization operate at the policy level, guiding reasoning trajectories, tool use, and verification. Finally, non-canonical chart understanding extends these paradigms to structurally diverse visualizations.

Because existing methods are evaluated on different benchmarks, chart types, and metrics, their raw scores are not always directly comparable. Instead, Tab. \ref{tab:fusion_comparison} provides a high-level synthesis of robust performance trends across paradigms, emphasizing which fusion strategy is best suited to which chart type, task difficulty, and deployment constraint.

\begin{table*}[t]
    \scriptsize
    \centering
    \setlength{\tabcolsep}{3pt}
    \renewcommand{\arraystretch}{1.45}
    \begin{tabular}{@{}>{\raggedright\arraybackslash}p{0.14\linewidth}>{\raggedright\arraybackslash}p{0.13\linewidth}>{\raggedright\arraybackslash}p{0.17\linewidth}>{\raggedright\arraybackslash}p{0.11\linewidth}>{\raggedright\arraybackslash}p{0.25\linewidth}>{\raggedright\arraybackslash}p{0.14\linewidth}@{}}
        \hline
        Paradigm & Fusion Level & Best-Fit Chart Types / Tasks & Task Difficulty & Comparative Trend & Cost and Scalability \\
        \hline
        Classical CV/DL and hybrid pipelines & Feature / decision fusion & Clean canonical charts; classification, element extraction, and factoid QA & Low & Efficient and interpretable for narrow tasks, but brittle under open-ended questions, diverse layouts, and unseen labels & Low inference cost, but high engineering cost for new chart types \\
        Component-assembled VLMs \cite{kim2021donut, lee2023pix2struct, liu2022matcha, masry2023unichart} & Architecture-level vision-language connection & Canonical statistical charts; chart derendering, captioning, and lightweight QA & Medium & Stronger than task-specific pipelines on unified chart-to-X tasks, but limited by shallow alignment between separately trained visual encoders and language decoders & Moderate training and inference cost; limited generalization to complex reasoning \\
        Adapted general-purpose MLLMs \cite{han2023chartllama, meng2024chartassisstant, masry2024chartgemma} & Parameter-level alignment through SFT or KD & Canonical and semi-structured charts; open-ended QA, summarization, and semantic analysis & Medium to high & Improves instruction following, prior knowledge, and open-ended generation; remaining errors often come from chart-specific grounding, OCR, and visual arithmetic & High fine-tuning cost; inference cost follows the backbone MLLM size \\
        Prompting and inference-time reasoning \cite{do2023llmsworkchartsdesigning, zhang2024tinychart, liu2024chartthinkercontextualchainofthoughtapproach} & Reasoning-level fusion & Charts with reliable visual perception; numerical, symbolic, and compositional QA & Medium to high & Low-cost gains on reasoning-heavy tasks by making intermediate steps explicit, but cannot repair misread axes, legends, or values & No training cost; higher token and latency cost for long reasoning chains \\
        Intermediate representations \cite{liu2022deplot, zhao2025chartcoder, ji2025socratic, cui2025draw} & Representation-level fusion & Numerical QA, chart-to-table, chart-to-code, SVG/XML reconstruction, and verifiable analysis & Medium to high & Often improves value grounding and executable verification by exposing structured evidence; performance depends on the fidelity of the generated representation & Extra conversion, parsing, or tool-execution cost; errors may propagate downstream \\
        Dynamic MoE and representation routing \cite{xu2024chartmoe, yang2024askchartuniversalchartunderstanding} & Architecture-level adaptive fusion & Heterogeneous canonical tasks requiring tables, code, JSON, or raw visual embeddings & High & Better matches representation choice to query type, e.g., tables for numerical reasoning and code for reconstruction; provides a clearer fusion taxonomy than single-representation models & Higher training complexity and routing overhead; requires expert specialization data \\
        RL, preference optimization, and visual tool policy \cite{jia2025chartreasonercodedrivenmodalitybridging, zhang2025enhancingcharttocodegenerationmultimodal, huang2025visualtoolagent, su2025openthinkimglearningthinkimages} & Policy / decision-level fusion & Long-chain reasoning, tool invocation, chart-to-code preference learning, and self-verification & High & Encourages exploration of reasoning paths and tool-use trajectories beyond static imitation; promising for hallucination reduction and multi-step analysis & Very high training cost; longer inference when multiple samples, tools, or verification loops are used \\
        Non-canonical structured and hybrid systems \cite{ye2024endtoendvlmsleveragingintermediate, omasa2025arrow, zhu2024multichartqa, li2025chartgalaxydatasetinfographicchart} & Structural and symbolic fusion & Flowcharts, infographics, scientific figures, maps, and multi-chart layouts & High & Best suited when topology recovery, region grounding, and cross-region synthesis are required; direct image-to-text generation alone is usually insufficient & Cost varies by pipeline; scalability is constrained by annotation, layout diversity, and tool robustness \\
        \hline
    \end{tabular}
    \caption{High-level comparison of fusion paradigms for chart understanding. The table summarizes cross-paper trends rather than directly comparing raw benchmark scores.}
    \label{tab:fusion_comparison}
\end{table*}

\begin{figure}
    \centering
    \includegraphics[width=\linewidth]{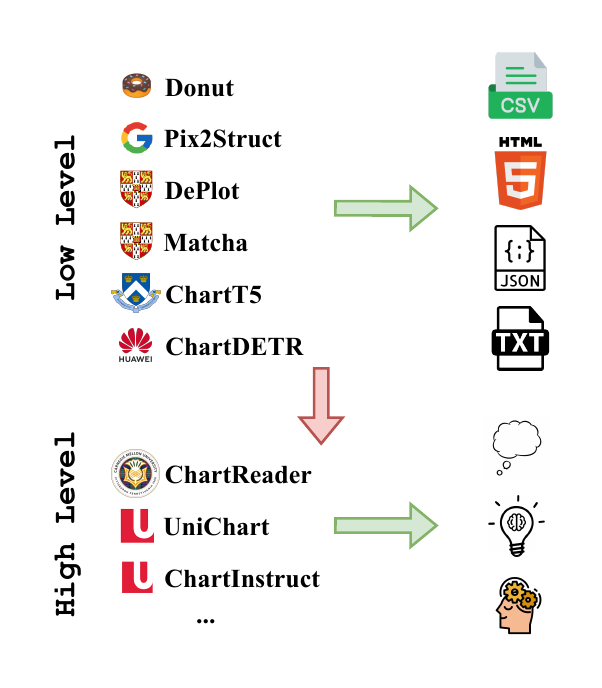}
    \caption{Two stages of component-assembled VLMs for chart understanding. Low-level models mainly convert chart images into structured representations such as tables, JSON, HTML, text, or code, whereas high-level models extend this pipeline toward semantic interpretation, chart captioning, and open-ended question answering.}
    \label{fig: early vlms}
\end{figure}

\subsubsection{Component-Assembled VLMs}
Component-assembled VLMs mark the first post-MLLM paradigm for chart understanding, which typically contain a Vision Encoder, a Language Decoder, and a vision-language connector. The Vision Encoder extracts textual and graphical signals from chart images, while the connector projects these features into the input space of the Language Decoder. Tab. \ref{tab: earlyVLMs} summarizes representative component-assembled VLMs.

\renewcommand{\arraystretch}{3}
\begin{table*}
    \footnotesize
    \centering
    \begin{tabular}{c c c c c}
        \hline
         Model & Parameter & Training Data & Architecture & Purpose \\
         \hline
         Donut \cite{kim2021donut} & 143M & IIT-CDIP \cite{10.1145/1148170.1148307}, SynthDoG & Swin Trans. \cite{liu2021swin} + BART \cite{lewis2019bart} & \makecell{Convert document \\ images into JSON.} \\
         Pix2Struct \cite{lee2023pix2struct} & \makecell{Base 282M, \\ Large 13B} & \makecell{Screenshot-HTML pairs, \\ BooksCorpus} & ViT \cite{dosovitskiy2020image} + Transformer Dec. & \makecell{Parse masked screenshots \\ of web pages into HTML.} \\
         DePlot \cite{liu2022deplot} & Undisclosed & \makecell{Web-Crawled and synthetic \\ plot-table pairs} & Pix2Struct \cite{lee2023pix2struct} & \makecell{Translate chart images \\ to linearized tables.} \\
         MatCha \cite{liu2022matcha} & \makecell{Base 282M, \\ Large 13B} & \makecell{Wikipedia, OWID, \\ OECD, Statista, etc.} & Pix2Struct \cite{lee2023pix2struct} & \makecell{Focus on chart derendering, \\ and math reasoning.} \\
        ChartReader \cite{cheng2023chartreader} & Undisclosed & \makecell{EC400K \cite{luo2021chartocr}, ChartQA \cite{masry2022chartqa}, \\ PlotQA \cite{methani2020plotqa}, etc.} & T5-Like Architecture & \makecell{A comprehensive chart- \\ understanding system \\ capable of C2T/CAP/QA.} \\
        ChartT5 \cite{zhou2023enhanced} & Undisclosed & \makecell{PlotQA \cite{methani2020plotqa}, DVQA \cite{kafle2018dvqa}, etc.} & VLT5 \cite{cho2021unifying} & Convert charts into tables. \\
         Unichart \cite{masry2023unichart} & 201M & \makecell{OWID, Statista, \\ OECD, PlotQA, etc.} & Donut \cite{kim2021donut}+ BART \cite{lewis2019bart} & \makecell{A comprehensive chart- \\ understanding system \\ capable of C2T/CAP/QA.} \\
        ChartDETR \cite{xue2023chartdetrmultishapedetectionnetwork} & Undisclosed & \makecell{Adobe Synthetic, \\ ExcelChart400k \cite{luo2021chartocr}} & CNN + Transformer & Target chart elements. \\
        ChartInstruct \cite{masry2024chartinstruct} & Undisclosed & \makecell{Unichart \cite{masry2023unichart}, \\ OECD, OWID} & \makecell{UniChart Vis. Enc. \cite{masry2023unichart} + \\ Llama2 \cite{touvron2023llamaopenefficientfoundation}/Flan-T5-XL \cite{chung2022scalinginstructionfinetunedlanguagemodels}} & \makecell{A unified model for chart \\ understanding and \\ reasoning.} \\
         \hline
    \end{tabular}
    \caption{An overview of the component-assembled VLMs for chart understanding. During this phase, the models were small in the parameter size, and were mostly assembled with components from different models.}
    \label{tab: earlyVLMs}
\end{table*}

\textbf{Low-Level Stage.} Early component-assembled models primarily focused on derendering and structured extraction. Donut \cite{kim2021donut} combines a Swin Transformer \cite{liu2021swin} Vision Encoder with multilingual BART \cite{lewis2019bart, liu2020multilingual} to map document images into JSON-format outputs. Pix2Struct \cite{lee2023pix2struct} further emphasizes visually situated language by processing screenshots as a single visual channel and using a variable-resolution Vision Transformer with a Transformer decoder \cite{vaswani2017attention} to reduce aspect-ratio distortion. Its screenshot-HTML pretraining helps the model learn spatial relationships between layout and text before adaptation to chart tasks.

Subsequent work made the task boundary more explicit: chart understanding requires both layout analysis and numerical reasoning. Built on Pix2Struct, MatCha \cite{liu2022matcha} uses chart derendering and math reasoning as complementary objectives, where derendering reconstructs tables or code and math reasoning trains the model to answer rendered mathematical questions. ChartT5 \cite{zhou2023enhanced} adapts VLT5 \cite{cho2021unifying} with a pre-trained Mask R-CNN object detector \cite{he2017mask} and introduces Masked Header Prediction (MHP) and Masked Value Prediction (MVP) to strengthen table-oriented cross-modal alignment.

\textbf{High-Level Stage.} While the aforementioned models excel at low-level data extraction (derendering), recent research focuses on unified frameworks capable of high-level semantic tasks, such as chart captioning and open-ended question answering.

ChartReader \cite{cheng2023chartreader} integrates a rule-free component detection module with a T5-like encoder-decoder architecture. Using an Hourglass network \cite{newell2016stacked}, it identifies key chart elements, feeds the fused features into the language model, and formulates chart-to-X tasks as unified QA tasks. In contrast, UniChart \cite{masry2023unichart} removes explicit object detection modules and builds an end-to-end framework on the Donut Vision Encoder \cite{kim2021donut} and BART \cite{lewis2019bart}. With task-specific prompts, a single model can support visual element extraction, numerical reasoning, and summarization.

ChartInstruct \cite{masry2024chartinstruct} further shows that high-level chart reasoning is also constrained by the capacity of the Language Decoder. By leveraging instruction tuning with stronger LLMs such as Flan-T5-XL \cite{chung2024scaling} and Llama-2 \cite{touvron2023llama2}, it improves semantic reasoning beyond low-level extraction. This observation motivates the next paradigm: adapting already aligned general-purpose MLLMs rather than assembling weaker components from scratch.

\paragraph{Insight.} Component-assembled VLMs mark a paradigm shift from task-specific chart pipelines toward vision-language modeling. By connecting visual encoders and language decoders with chart-oriented training objectives, they are effective for derendering and lightweight reasoning over clean canonical charts. However, their fusion remains limited for complex reasoning, open-ended analysis, and non-canonical layouts, which motivates later work on adapting stronger general-purpose MLLMs with chart-specific alignment data.


\begin{figure*}
    \centering
    \includegraphics[width=\linewidth]{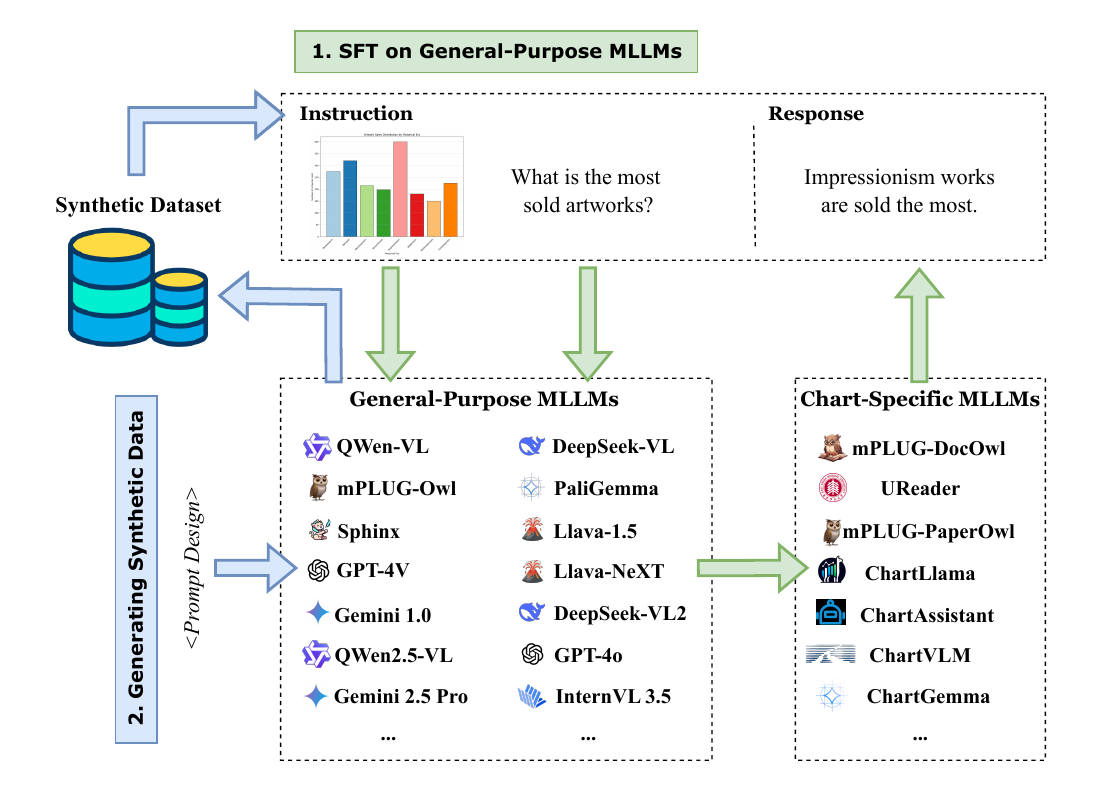}
    \caption{Adapting general-purpose MLLMs. The first usage is to fine-tune these large-scale and well-pretrained MLLMs to adapt themselves to the scenarios in Chart Understanding. The second usage is to take advantage of the abundant pre-trained knowledge of these general-purpose MLLMs and create high-quality synthetic datasets.}
    \label{fig: sft}
\end{figure*}

\renewcommand{\arraystretch}{3.5}
\begin{table*}
    \scriptsize
    \centering
    \begin{tabular}{c c c c c}
        \hline
         Model & Original Model & Purpose & Training Data & Fine-Tuning Detail \\
         \hline
         mPLUG-DocOwl \cite{ye2023mplug} & mPLUG-Owl \cite{ye2024mplugowlmodularizationempowerslarge} & General QA & \makecell{ChartQA \cite{masry2022chartqa}, KLC \cite{Stanis_awek_2021}, \\ InfographicVQA \cite{mathew2021infographicvqa}, WTQ \cite{pasupat2015compositionalsemanticparsingsemistructured}, \\ TabFact \cite{chen2020tabfactlargescaledatasettablebased}, TextCaps \cite{sidorov2020textcapsdatasetimagecaptioning}, \\ DocVQA \cite{mathew2021docvqa}, etc.} & \makecell[l]{Stage 1: Fine-tune the visual abstractor and \\ LoRA in the language decoder for 10 epochs. \\ Stage 2: Only fine-tune the language decoder \\ for 3 epochs.} \\
         UReader \cite{ye2023ureaderuniversalocrfreevisuallysituated} & mPLUG-Owl \cite{ye2024mplugowlmodularizationempowerslarge} & General QA & \makecell{VisualMRC \cite{tanaka2021visualmrcmachinereadingcomprehension}, DocVQA \cite{mathew2021docvqa}, \\ WTQ \cite{pasupat2015compositionalsemanticparsingsemistructured}, TabFact \cite{chen2020tabfactlargescaledatasettablebased}, \\ TextVQA \cite{singh2019vqamodelsread}, ChartQA \cite{masry2022chartqa}, etc.} & \makecell[l]{Fine-tune the visual abstractor, crop position \\ encoding, and LoRA in the language decoder \\ for 10 epochs.} \\
         mPLUG-PaperOwl \cite{hu2024mplug} & mPLUG-DocOwl \cite{ye2023mplug} & General QA & M-Paper \cite{hu2024mplug} & \makecell[l]{Fine-tune the visual abstractor and LoRA in \\ the language decoder for 10 epochs.} \\
         ChartLlama \cite{han2023chartllama} & Llava-1.5 \cite{liu2024improvedbaselinesvisualinstruction} & General QA & Data Generated by GPT-4 & \makecell[l]{Fine-tune the projection layer and LoRA in \\ the LLM.} \\
         ChartAssistant \cite{masry2024chartinstruct} & \makecell{ChartAst-D: Donut, \\ ChartAst-S: Sphinx} & General QA & ChartSFT \cite{masry2024chartinstruct} & \makecell[l]{Stage 1: Pre-train on Chart-to-Table Trans- \\ lation. \\ Stage 2: Train the model with ChartSFT.} \\
        ChartVLM \cite{xia2024chartx} & \makecell{Pix2Struct \cite{lee2023pix2struct} + \\ Vicuna \cite{chiang2023vicuna}} & General QA & \makecell{ChartQA \cite{masry2022chartqa}, PlotQA \cite{methani2020plotqa}, \\ SimChart9K \cite{xia2024structchartschemametricaugmentation}, etc.} & \makecell[l]{Fine-tune all the weights in Pix2Struct and \\ the LoRA in Vicuna.} \\
        ChartGemma \cite{masry2024chartgemma} & PaliGemma \cite{beyer2024paligemma} & General QA & \makecell{PlotQA \cite{methani2020plotqa}, WebCharts \cite{masry2024chartinstruct}, \\ ChartFC \cite{akhtar2023readingreasoningchartimages}, etc.} & \makecell[l]{Freeze the vision encoder, and only fine-tune \\ the language decoder.}\\
         \hline
    \end{tabular}
    \caption{An overview of the fine-tuned models. In this phase, the models were no longer assembled with components from different models. They were mostly developed from general-purpose models through SFT.}
    \label{tab: sft}
\end{table*}

\subsubsection{Adapting General-Purpose MLLMs}
General-purpose MLLMs address two weaknesses of component-assembled VLMs: insufficient cross-modal alignment among heterogeneous modules and limited language-side reasoning capacity. Since these models are pretrained on substantially broader multimodal corpora, they provide stronger instruction following, prior knowledge, and zero/few-shot reasoning. Their remaining weakness is chart-specific grounding, especially fine-grained OCR, axis-value alignment, and numerical perception. As shown in Fig. \ref{fig: sft}, recent work mainly adapts these models through Parameter-Efficient Fine-Tuning (PEFT) via Supervised Fine-Tuning (SFT), and Knowledge Distillation (KD) via synthetic data generation. Tab. \ref{tab: sft} summarizes representative fine-tuned models.

\textbf{Parameter-Efficient Fine-Tuning via Supervised Fine-Tuning.} Directly applying standard MLLMs to charts is hindered by resolution loss and domain mismatch. UReader \cite{ye2023ureaderuniversalocrfreevisuallysituated} addresses high-resolution inputs with a Shape-Adaptive Cropping module that fuses local crops with a low-resolution global view, and uses LoRA \cite{hu2022lora} to update only a small fraction of parameters. Similarly, mPLUG-DocOwl \cite{ye2023mplug} aligns the visual abstractor with the LLM through a two-stage training paradigm, emphasizing instruction tuning for visually situated document understanding.

Some models extend this adaptation beyond isolated chart images. mPLUG-PaperOwl \cite{hu2024mplug} incorporates captions and surrounding paper text, allowing chart-level visual features to be interpreted with document-level context. ChartAssistant \cite{meng2024chartassisstant} explores the scalability of SFT by training ChartAst-D and ChartAst-S on extensive instruction pairs, showing that domain-specific instructions can outperform disjoint component assembly. Overall, SFT and PEFT are most suitable when chart-specific instruction data are available and the target model already has strong general multimodal alignment; their main trade-off is the training cost and the risk of overfitting to narrow chart distributions or annotation styles.

\textbf{Knowledge Distillation via Synthetic Data Generation.} The scarcity of high-quality, annotated chart-instruction pairs remains a bottleneck for SFT. To address this, researchers leverage state-of-the-art proprietary models like GPT-4 to synthesize training data, effectively distilling the powerful teacher model's knowledge into smaller, open-source models.

ChartLlama \cite{han2023chartllama} uses GPT-4 to generate Python code, render diverse charts, derive ground-truth tables, and produce instruction-tuning QA pairs. This synthetic corpus allows ChartLlama, adapted from LLaVA-1.5 \cite{liu2023visual, liu2024improvedbaselinesvisualinstruction}, to learn from controlled chart metadata rather than noisy real-world annotation alone. 

ChartX \cite{xia2024chartx} and CHOPINLLM \cite{fan2024pre} extend this data-centric paradigm to benchmarks with varying cognitive complexity. ChartX introduces ChartVLM, where a base decoder \cite{lee2023pix2struct} handles low-level extraction and an auxiliary LLM \cite{chiang2023vicuna} is conditionally activated for higher-level reasoning, matching computation to task complexity.

Because metadata-driven synthesis may bias models toward textual tables rather than visual semantics, ChartGemma \cite{masry2024chartgemma} generates instructions directly from in-the-wild chart images without accessing source tables. Using Gemini 1.5 Flash \cite{team2023gemini} as the teacher, it constructs visual instruction data that is less dependent on table-to-chart rendering pipelines. KD is therefore most useful when human annotations are scarce but a strong teacher can provide diverse supervision; its trade-off is that teacher hallucinations, synthetic rendering artifacts, or hidden table priors may be transferred to the student model.

\paragraph{Insight.} Adapting general-purpose MLLMs alleviates two major limitations of component-assembled VLMs: weak vision-language alignment caused by heterogeneous components, and insufficient reasoning ability or prior knowledge in smaller language backbones. By starting from well-aligned and instruction-following MLLMs, this paradigm better supports open-ended chart question answering and higher-level semantic analysis. However, chart-specific fine-tuning of large MLLMs remains computationally expensive, while these models already possess strong in-context learning abilities. This motivates lower-cost prompting and inference-time strategies that adapt MLLMs to chart understanding without updating model parameters.

\definecolor{myyellow}{RGB}{200, 182, 85}
\definecolor{myred}{RGB}{183, 52, 52}
\definecolor{myblue}{RGB}{67, 145, 219}
\definecolor{mygreen}{RGB}{35, 144, 37}
\definecolor{mypurple}{RGB}{137, 53, 182}

\renewcommand{\arraystretch}{3}
\begin{table*}
    \scriptsize
    \centering
    \begin{tabular}{llll}
        \hline
        \makecell{Specific Task} & \makecell[l]{CCR Prompt Format} & Question & Answer\\
        \hline
        \makecell[l]{Visual Retrieval} & \makecell[l]{No need for CCR} & - & - \\
        \makecell[l]{Numerical Retrieval} & \makecell[l]{No need for CCR} & - & - \\
        \makecell[l]{Compositional Retrieval} & \makecell[l]{No need for CCR} & - & - \\
        
        \makecell[l]{Complex Retrieval} & \makecell[l]{\textcolor{myyellow}{operands with visual attributes}, \\ \textcolor{myred}{operands without visual attributes}, \\ \textcolor{mygreen}{reasoning}, \textcolor{mypurple}{result}} & \makecell[l]{How many data points on the \\ disapprove line are above 50?} & \makecell[l]{\textcolor{myred}{The disapprove number 51 and 50} \\ \textcolor{mygreen}{are greater than 50}. \textcolor{mypurple}{The answer is 2}.} \\
        
        \makecell[l]{Calculation Reasoning} & \makecell[l]{\textcolor{myred}{operands without visual attributes}, \\ \textcolor{myblue}{operators}, \textcolor{mygreen}{reasoning}, \textcolor{mypurple}{result}} & \makecell[l]{What’s the ratio of Lean Republican \\ segment and Republican segment?} & \makecell[l]{\textcolor{myred}{The value of Lean Republican segment is 39,} \\ \textcolor{myred}{and the value of Republican segment is 53}. \\ \textcolor{myblue}{The ratio is 39 / 53 = 0.7358}, \textcolor{mypurple}{ the answer}.}\\
        
        \makecell[l]{Visual Reasoning} & \makecell[l]{\textcolor{myyellow}{operands with visual attributes}, \\ \textcolor{myblue}{operators}, \textcolor{mygreen}{reasoning}, \textcolor{mypurple}{result}} & \makecell[l]{What’s the average value of the \\ first two blue bars in the chart?} & \makecell[l]{\textcolor{myyellow}{The first two blue bars are 30-34 years old with} \\ \textcolor{myyellow}{0.35 and 25-29 years old with 0.31}. \textcolor{myblue}{The average} \\ \textcolor{myblue}{value is ( 0.35 + 0.31 ) / 2 = 0.33}, \textcolor{mypurple}{the answer}.}\\
        
        \makecell[l]{Compositional Reasoning} & \makecell[l]{\textcolor{myyellow}{operands with visual attributes}, \\ \textcolor{myred}{operands without visual attributes}, \\ \textcolor{myblue}{operators}, \textcolor{mygreen}{reasoning}, \textcolor{mypurple}{result}} & \makecell[l]{Is the average value of Andean Latin \\ America and Cambodia more than \\ the value of Thailand?} & \makecell[l]{\textcolor{myred}{The values of Andean Latin America, Cambodia}  \\ \textcolor{myred}{and Thailand are 1.47, 0.77 and 0.39.} \textcolor{myblue}{The aver.} \\ \textcolor{myblue}{of Andean Latin America and Cambodia is 1.12}. \\ \textcolor{mygreen}{Since, 1.12 > 0.39}, \textcolor{mypurple}{the answer is yes}.} \\
        
        \hline
    \end{tabular}
    \caption{The CCR format for different types of questions. We integrate "Add \& Subtraction" and "Division \& Multiplication" tasks in the original paper \cite{do2023llms} into "Calculation Reasoning", and "All Types of Reasoning" into "Compositional Reasoning". Simple retrieval queries require no CCR prompts, while complex retrieval queries and reasoning queries require CCR prompts to break questions into atomic reasoning steps.}
    \label{tab: ccr}    
\end{table*}

\subsubsection{Prompting and Inference-Time Reasoning}
Prompting and inference-time reasoning adapt general-purpose MLLMs without changing model parameters. This paradigm focuses on \emph{how} the model organizes its reasoning after visual features have already been encoded. It is therefore distinct from intermediate-representation methods in Sec. \ref{sec: augmentation}, which modify \emph{what} visual information is exposed to the language model.

\textbf{Chain of Chart Reasoning (CCR).} PROMPTCHART \cite{do2023llmsworkchartsdesigning} pioneered few-shot prompting for Chart Understanding without gradient updates. It combines a Visual Data Table Generator, a Prompt Constructor, and InstructGPT \cite{ouyang2022traininglanguagemodelsfollow}. Its Chain of Chart Reasoning (CCR) decomposes complex queries into atomic steps, as shown in Tab. \ref{tab: ccr}, by explicitly specifying operands, operators, reasoning, and final results. In this sense, CCR improves reasoning structure rather than visual encoding itself. It is most suitable for complex retrieval or compositional QA where relevant operands can already be identified, but it remains fragile when the upstream table generator or visual reader extracts incorrect values.

\begin{figure*}
    \centering
    \includegraphics[width=\linewidth]{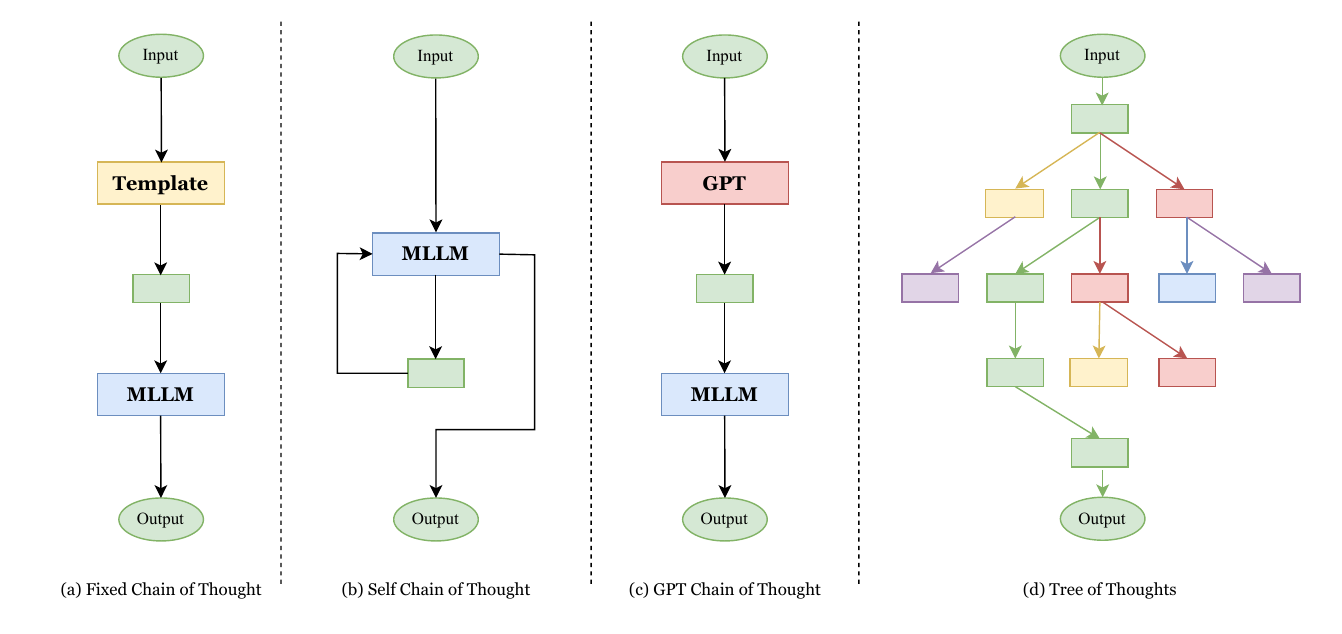}
    \caption{A schematic comparison of prompting-based reasoning strategies for chart understanding, including Fixed CoT, Self-CoT, GPT-CoT, and Tree of Thoughts.}
    \label{fig: prompt}
\end{figure*}

\textbf{Evolving Chain-of-Thought (CoT) Strategies.} CoT prompting further varies by how the reasoning chain is obtained. As shown in Fig. \ref{fig: prompt}, current strategies include Fixed CoT, where templates are manually defined; Self-CoT, where the model generates its own reasoning path; and GPT-CoT, where a stronger teacher model such as GPT-4 generates reasoning demonstrations. Results on ChartBench \cite{xu2023chartbench} suggest that GPT-CoT is effective for distilling reasoning into smaller models (e.g., MiniGPT-v2 \cite{chen2023minigptv2largelanguagemodel}, Qwen-VL \cite{bai2023qwen}, Internlm-XComposer-v2 \cite{dong2024internlm}), while Self-CoT may introduce hallucinated intermediate steps. These strategies are suitable for multi-step reasoning when visual grounding is already reliable; the trade-off is between template rigidity in Fixed CoT, teacher cost in GPT-CoT, and self-generated reasoning errors in Self-CoT.

\textbf{Program-of-Thoughts (PoT) and Symbolic Fusion.} PoT targets numerical reasoning rather than visual reconstruction. TinyChart \cite{zhang2024tinychart} integrates Program-of-Thoughts (PoT) \cite{chen2022program, gao2023palprogramaidedlanguagemodels} into SFT, uses Visual Token Merging for efficient high-resolution encoding, and trains the Language Decoder to generate executable Python code for computational queries. This differs from chart-to-code representation methods in Sec. \ref{sec: augmentation}: here, code is a reasoning trace for arithmetic, not necessarily a faithful reconstruction of the whole chart. PoT is most suitable for arithmetic-intensive questions after the relevant chart values have been localized, but it cannot compensate for wrong value extraction and may introduce execution or syntax-related overhead.

\textbf{Non-Linear Reasoning and Contextual Fusion.} Beyond linear chains, ChartThinker \cite{liu2024chartthinkercontextualchainofthoughtapproach} introduces a Context-Enhanced Chain-of-Thought (C-CoT) approach related to Tree of Thoughts (ToT) \cite{yao2023treethoughtsdeliberateproblem}, as shown in Fig. \ref{fig: prompt}. It retrieves k-nearest chart-caption pairs as in-context exemplars, explores multiple reasoning paths, and fuses their conclusions at the decision level. This improves robustness for open-ended reasoning, but still assumes that the underlying visual evidence has been encoded with sufficient fidelity. Non-linear reasoning is most appropriate for ambiguous, open-ended, or multi-hop questions where a single linear chain is insufficient; its cost is higher latency, more tokens, and a harder final path-selection problem.

\paragraph{Insight.} Prompting and inference-time reasoning provide a low-cost way to transfer the capabilities of general-purpose MLLMs to chart understanding without updating model parameters. CCR, CoT, PoT, and ToT can make symbolic, numerical, and compositional reasoning more explicit, thereby improving the use of the model's existing reasoning ability. However, prompt engineering is still constrained by the model's own perceptual bottleneck: dense pixel information must be compressed into semantic token space, where fine-grained chart details can be lost. Therefore, prompting alone cannot fundamentally solve weak visual-textual alignment, motivating intermediate representations and modality alignment methods that expose chart structure more explicitly.

\begin{figure*}
    \centering
    \includegraphics[width=\linewidth]{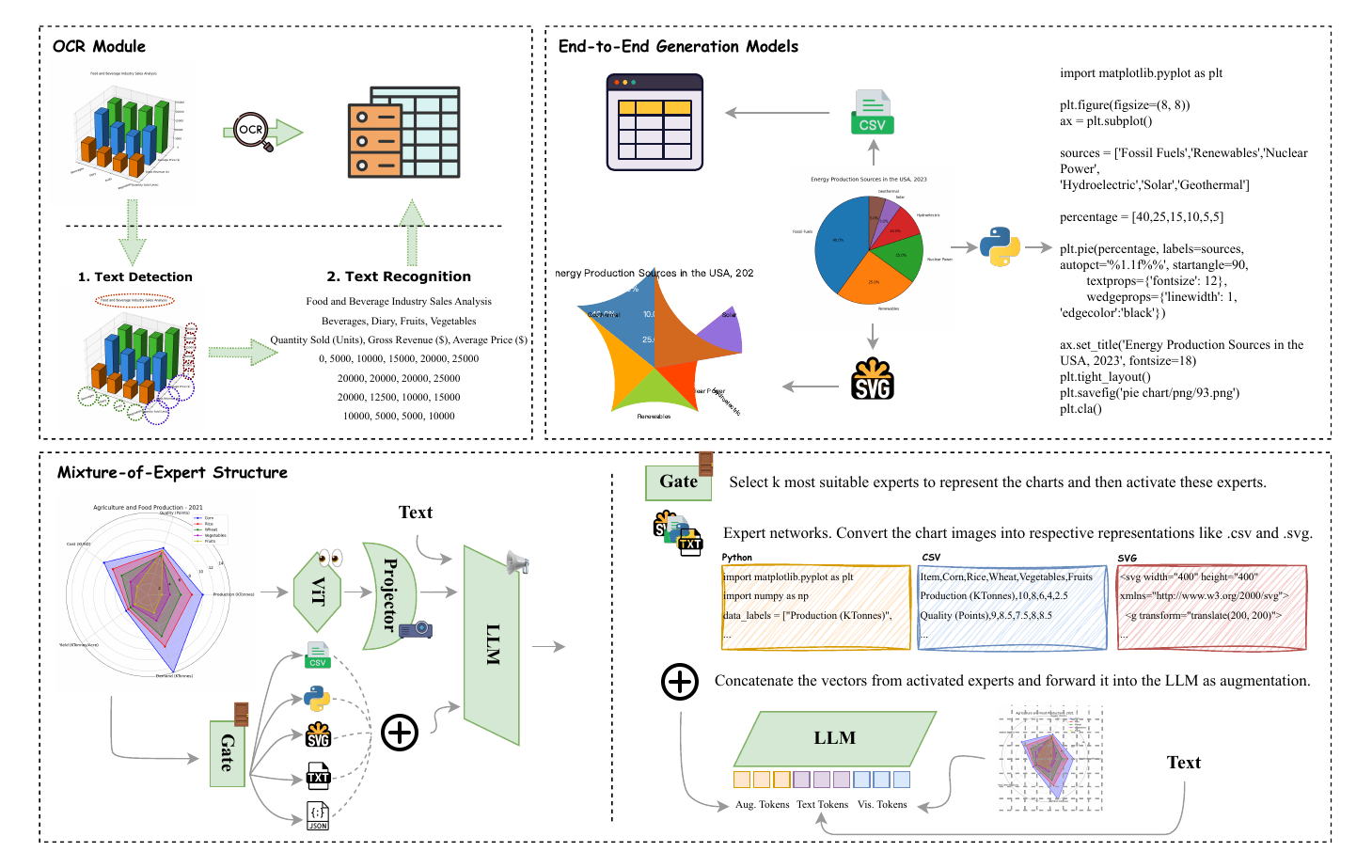}
    \caption{Three stages in the development of augmentation tools. The first stage is the OCR modules. They usually separate the whole process into two phases, each of which locates the text regions and recognizes the text, respectively. The second stage is end-to-end generation models. They use a unified and end-to-end architecture design, enabling the models to learn different levels and different forms of representations. The third stage is mixture-of-experts (MoE), which enables the models to learn multiple forms of representations within a single architecture.}
    \label{fig: augmentation}
\end{figure*}

\subsubsection{Intermediate Representation and Modality Alignment} \label{sec: augmentation}
A primary challenge in MLLMs is the modality gap between dense chart pixels and the semantic token space of LLMs. Unlike prompting, which reorganizes reasoning at inference time, intermediate representation methods change the input evidence itself by converting chart images into explicit symbolic or structural forms. Their goal is to preserve numerical values, layout, and visual relations that may be lost in raw visual embeddings.

\textbf{From OCR to End-to-End Derendering.} As shown in Fig. \ref{fig: augmentation}, early approaches \cite{liu2016star, shi2016robust, shi2016end, wang2017gated, cheng2017focusing, borisyuk2018rosetta, luo2021chartocr, yang2024askchartuniversalchartunderstanding} relied on OCR modules to extract textual elements in two-stage detection-recognition pipelines. These systems are useful for labels and titles, but they lack holistic chart understanding and can fail on complex layouts or overlapping elements.

End-to-end generation models such as Donut \cite{kim2021donut}, DEPLOT \cite{liu2022deplot}, and Pix2Struct \cite{lee2023pix2struct} reduce this dependence on cascaded OCR by mapping whole chart images into token sequences. They implicitly learn spatial relationships, but the generated sequence must still choose a representation format. This leads to three major families: tables, programmatic code, and vector graphics.

\textbf{Symbolic and Executable Representations.} While end-to-end models can generate unstructured text, recent work emphasizes structured intermediate representations to facilitate better fusion with LLMs.

\begin{itemize}
    \item \textbf{Linearized Data Tables.} DEPLOT \cite{liu2022deplot} translates chart images into linearized textual tables, which can be directly used by LLMs for numerical QA and reasoning. Its Relative Mapping Similarity (RMS) metric evaluates whether the generated table preserves relative value relationships even when exact values are imperfect. SIMPLOT \cite{kim2025simplotenhancingchartquestion} further refines table generation through a Knowledge Distillation framework that trains a student encoder to suppress visual noise and produce cleaner tabular evidence. Tables are most suitable for value lookup, comparison, and aggregation over canonical charts, but they compress away layout, color, mark shape, and other visual semantics that may be essential for open-ended interpretation.
    
    \item \textbf{Programmatic Code.} Tables expose values but discard visual design. Programmatic representations instead encode both data and rendering choices through executable scripts \cite{xu2025improvediterativerefinementcharttocode}. ChartCoder \cite{zhao2025chartcoder, zhao2025chartcoderadvancingmultimodallarge}, built with SigLIP-384 \cite{zhai2023sigmoid} and DeepSeek Coder 6.7B \cite{guo2024deepseek}, uses a two-stage training paradigm: first aligning chart images with captions through the vision-language connector, then fine-tuning all components for chart-to-code translation. This representation is especially useful when visual fidelity and executable verification matter. Its trade-off is that code generation introduces syntax errors, execution cost, and possible mismatch between the rendered output and the original chart.
    
    \item \textbf{Vector Graphics (SVG/XML).} SVG/XML representations preserve geometric and hierarchical structure more explicitly than tables or plain text \cite{chen2023mystiquedeconstructingsvgcharts, moured2024chartformerlargevisionlanguage, ji2025socratic, chen2025visanatomysvgchartcorpus, shiinoki2025overcoming, cui2025draw}. Socratic Chart \cite{ji2025socratic} uses multi-agent generation and critique to reconstruct high-fidelity SVG attributes, while Draw with Thought (DwT) \cite{cui2025draw} decomposes charts into symbolic abstractions and generates editable mxGraph XML verified by Draw.io. These methods are closer to structural reconstruction than prompting-based reasoning. SVG/XML is suitable when geometry, hierarchy, and editability are central, but long structured sequences are difficult to generate faithfully and are especially brittle for noisy or highly non-canonical layouts.
\end{itemize}

\textbf{Dynamic Fusion via Mixture of Experts.} The above methods usually rely on one representation format. However, different questions may require different evidence: tables for values, code for reconstruction, and visual embeddings for style or layout. MoE-based methods therefore introduce architecture-level routing to dynamically combine representations \cite{jacobs1991adaptive, shazeer2017outrageously, zoph2022stmoedesigningstabletransferable}.

As shown in Fig. \ref{fig: augmentation}, an MoE framework typically contains specialized experts, a gating network that selects which experts to activate, and an aggregation module that fuses the selected outputs. This differs from RL-based tool selection in the next subsection: MoE learns representation routing inside the model architecture, whereas RL optimizes action or tool-use policies through reward feedback.

There is an emerging number of works leveraging MoE in Chart Understanding \cite{xu2024chartmoe, yang2024askchartuniversalchartunderstanding}. Replacing the Multi-Layer Perceptron (MLP) in InternLM-XComposer-v2 \cite{dong2024internlm}, ChartMoE \cite{xu2024chartmoe} employs a dynamic gating network to route visual features to task-specific experts. There are four experts, pretrained with distinct alignment tasks respectively, such as chart-to-table, chart-to-JSON, chart-to-code, and the original visual embeddings. The training involves a two-stage protocol. The first stage is Expert Pre-Training, where experts are trained independently on their respective representations while Vision Encoder is frozen. The second stage is end-to-end Supervised Fine-Tuning, where the gating network is trained to select the top-k experts based on the input query. The outputs of activated experts are fused via weighted summation and forwarded to the LLM. This allows ChartMoE to dynamically adapt its representation strategy, using code for visualization tasks and tables for numerical reasoning within a single unified framework. MoE is most suitable for heterogeneous tasks where no single representation is consistently optimal, but it adds routing complexity, expert specialization requirements, and possible error propagation from incorrect expert selection.

\paragraph{Insight.} Intermediate representations compensate for the information loss that occurs when raw chart images are compressed into visual tokens. By converting charts into tables, code, SVG/XML, or other symbolic forms, they expose numerical values, visual structure, and spatial relations in formats that LLMs can manipulate more reliably. However, this explicit conversion also introduces additional computational cost, especially when multiple representations or tool calls are required. Moreover, accurately translating complex non-canonical charts into faithful intermediate representations remains challenging, since errors in the intermediate form can propagate into downstream reasoning.

\begin{figure}
    \centering
    \includegraphics[width=\linewidth]{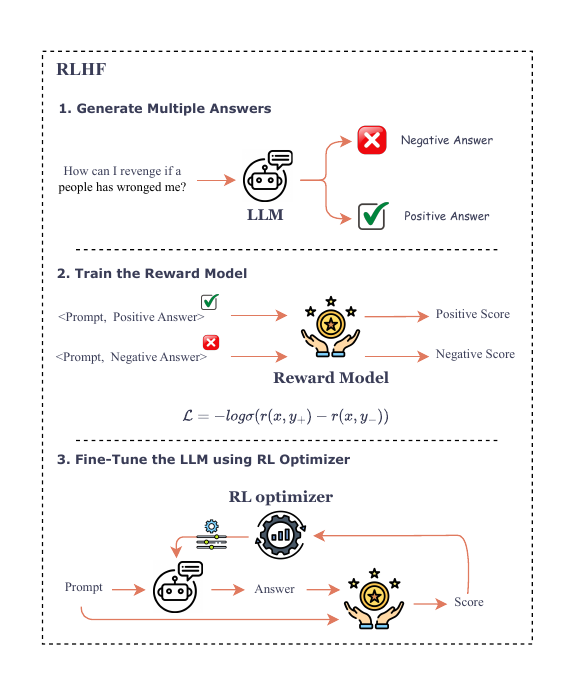}
    \caption{Pipeline of Reinforcement Learning from Human Feedback (RLHF), where preference-ranked responses are used to train a reward model and then optimize the policy model.}
    \label{fig: rlhf}
\end{figure}

\begin{figure}
    \centering
    \includegraphics[width=\linewidth]{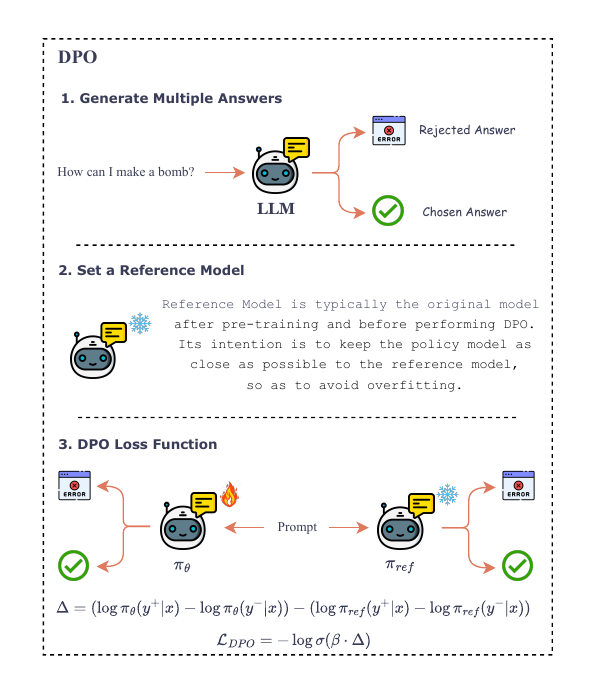}
    \caption{Pipeline of Direct Preference Optimization (DPO), which optimizes preference pairs directly against a reference model without training a separate reward model.}
    \label{fig: dpo}
\end{figure}

\subsubsection{Reinforcement Learning and Preference Optimization}
Reinforcement learning and preference optimization form a policy-level paradigm. Unlike SFT, which imitates labeled demonstrations, RL allows models to explore alternative outputs, reasoning paths, and tool-use trajectories under reward feedback. Recent Chart Understanding methods \cite{li2024insightableinsightdrivenhierarchicaltable, jia2025chartreasonercodedrivenmodalitybridging, zhang2025enhancingcharttocodegenerationmultimodal, su2025openthinkimglearningthinkimages, wu2025tabler1regionbasedreinforcementlearning, shi2025chartisttaskdriveneyemovement, huang2025visualtoolagent, zhang2025mllmsreallyunderstandcharts, huang2026sketchvlpolicyoptimizationfinegrained} therefore use RL to improve preference alignment, long-chain reasoning, and tool selection.

\textbf{Direct Preference Optimization.} Classical RLHF \cite{ouyang2022traininglanguagemodelsfollow, stiennon2022learningsummarizehumanfeedback} first collects multiple responses for the same prompt, ranks them according to human preference, and trains a reward model from these preference signals. The policy model is then optimized with PPO \cite{schulman2017proximalpolicyoptimizationalgorithms} under reward feedback, as shown in Fig. \ref{fig: rlhf}. DPO \cite{rafailov2024directpreferenceoptimizationlanguage}, illustrated in Fig. \ref{fig: dpo}, simplifies this pipeline by directly optimizing pairwise preferences while using a reference model to constrain the updated policy from drifting too far from the original model. In chart understanding, this is useful when correctness depends not only on the final text answer, but also on visual faithfulness or executable output quality. DPO is most suitable when pairwise preferences can be constructed reliably, but its effectiveness depends strongly on the coverage and quality of the preference pairs.

In chart-to-code generation, the Chart2Code pipeline \cite{zhang2025enhancingcharttocodegenerationmultimodal} applies iterative preference learning to improve both code correctness and visual fidelity. Starting from chart images and ground-truth plotting code, it generates candidate code, constructs flawed variants through perturbations over chart type, data, layout, color, text, and style, and scores candidates with both rendered-image evaluation and code-level evaluation \cite{yang2025chartmimicevaluatinglmmscrossmodal}. Preference pairs are retained only when the generated code outperforms the flawed variant on both evaluators, enabling iterative refinement toward visually faithful executable code. This design is suitable for outputs that can be compared by executable or rendered evidence, but it requires reliable evaluators and repeated candidate generation.

 \begin{figure*}
     \centering
     \includegraphics[width=1\linewidth]{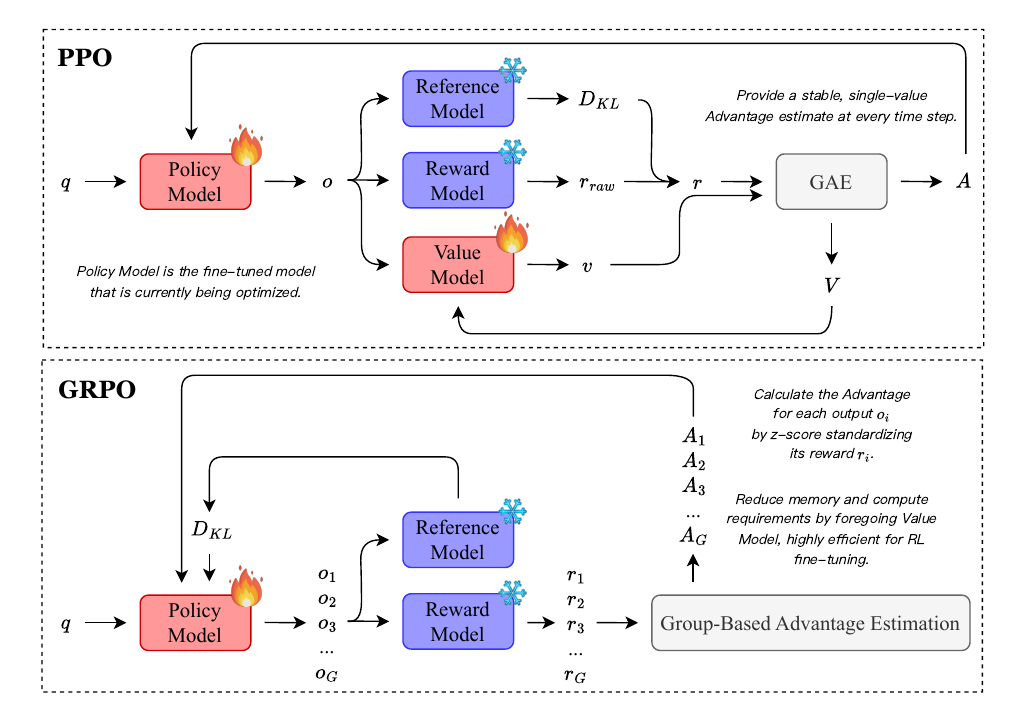}
     \caption{Comparison between PPO and GRPO. PPO relies on value estimation during policy optimization, whereas GRPO estimates relative advantages from a group of sampled outputs.}
     \label{fig: ppogrpo}
 \end{figure*}

\textbf{Long Chain Reasoning via Group Relative Policy Optimization.} DeepSeek-R1 \cite{deepseekai2025deepseekr1incentivizingreasoningcapability} shows that RL can incentivize long-chain reasoning with less dependence on supervised labels. For chart understanding, this matters because complex questions require multi-hop inference \cite{yao2023treethoughtsdeliberateproblem, Besta_2024}, tool use \cite{yao2023reactsynergizingreasoningacting}, and feedback-based refinement \cite{shinn2023reflexionlanguageagentsverbal}. This idea has been transferred to Vision-Language Models through R1-OneVision \cite{yang2025r1onevisionadvancinggeneralizedmultimodal} and Vision-R1 \cite{huang2025visionr1incentivizingreasoningcapability}, which combine SFT and RL to train models with image-question-answer data and long-chain reasoning traces. This paradigm is suitable when answer quality depends on intermediate reasoning behavior rather than final-answer imitation alone, but it is less reliable if rewards cannot distinguish grounded reasoning from plausible verbal explanations.

GRPO \cite{shao2024deepseekmathpushinglimitsmathematical}, shown in Fig. \ref{fig: ppogrpo}, is a critic-free alternative to PPO \cite{schulman2017proximalpolicyoptimizationalgorithms}. In both paradigms, the policy model generates candidate outputs, the reference model constrains policy drift through a KL penalty, and a reward function or reward model assigns scalar feedback to candidate responses. Unlike PPO, which relies on value estimation, GRPO samples a group of outputs for the same prompt and uses their relative rewards to estimate advantages. This makes it suitable for optimizing diverse reasoning paths, although it still requires repeated sampling and reward evaluation. Therefore, GRPO is particularly appropriate for long-chain chart reasoning and tool-use exploration, but its trade-off is higher inference-time sampling cost and strong dependence on well-designed reward signals.

ChartReasoner \cite{jia2025chartreasonercodedrivenmodalitybridging} adapts this methodology to fuse visual encoding with logical reasoning. The framework first converts raster chart images into semantically rich ECharts codes. Subsequently, the model is trained using a hybrid strategy. Initial SFT aligns the model with Chain of Thought data generated by reasoning LLMs like DeepSeek-R1 \cite{deepseekai2025deepseekr1incentivizingreasoningcapability}, establishing a foundational reasoning structure. To mitigate over-reasoning hallucinations, Chart Reasoner employs GRPO. By sampling multiple reasoning paths for a single chart query and scoring them based on answer correctness and conciseness, the model learns to favor efficient, logically sound reasoning trajectories over verbose or erroneous ones. Such code-driven GRPO is suitable for questions requiring both structural reconstruction and reasoning, but errors in the intermediate code can still constrain the reward signal.

\textbf{Visual Tool Selection via Group Relative Policy Optimization.} RL also clarifies the boundary between representation generation and tool-use policy. Sec. \ref{sec: augmentation} discussed representations such as tables, code, and SVG/XML; this subsection instead focuses on how a model learns when to invoke such tools. Prior tool-use methods either rely on large-scale SFT \cite{schick2023toolformerlanguagemodelsteach, liu2023llavapluslearningusetools} or training-free prompting \cite{gou2024toratoolintegratedreasoningagent, hu2024visualsketchpadsketchingvisual, lu2025octotoolsagenticframeworkextensible}, while RL directly optimizes dynamic decision making \cite{10.5555/3312046}.

VisTA \cite{huang2025visualtoolagent} uses GRPO to explore tool-invocation trajectories and rewards correct answers while discouraging unnecessary tool calls. OpenThinkImg \cite{su2025openthinkimglearningthinkimages} extends this idea by treating intermediate visual thoughts as tool-use actions and optimizing their selection with GRPO. These methods belong to decision-level fusion: the model learns which visual tools or intermediate aids to use for a query, rather than simply receiving a fixed representation. Tool-policy optimization is suitable when different questions require different external aids, but it increases inference latency and makes reward design more complex.

\paragraph{Insight.} Reinforcement learning and preference optimization move chart understanding from static imitation of labeled data toward dynamic exploration guided by rewards. This shift is especially useful for multi-step reasoning and external tool invocation, where the model must learn not only what answer to produce, but also which reasoning path or tool-use trajectory to follow. From an information fusion perspective, RL operates at the policy or decision level by optimizing how visual evidence, intermediate representations, and reasoning steps are selected and combined. However, RL-based methods usually require higher computational resources and longer inference time, making efficiency and reward design critical for practical deployment.

\begin{figure*}
    \centering
    \includegraphics[width=1\linewidth]{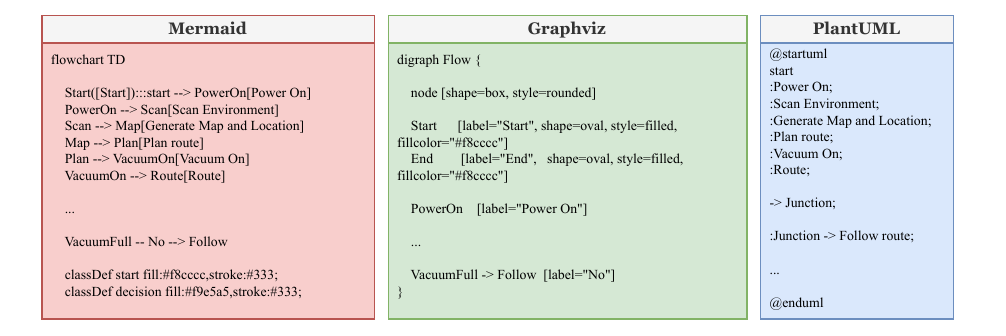}
    \caption{Graph description language, such as Mermaid, Graphviz, and PlantUML, is highly suitable for representing flowcharts, thus providing supplementary information for the Reasoner MLLM.}
    \label{fig: flowchart}
\end{figure*}

\begin{figure*}
    \centering
    \includegraphics[width=\linewidth]{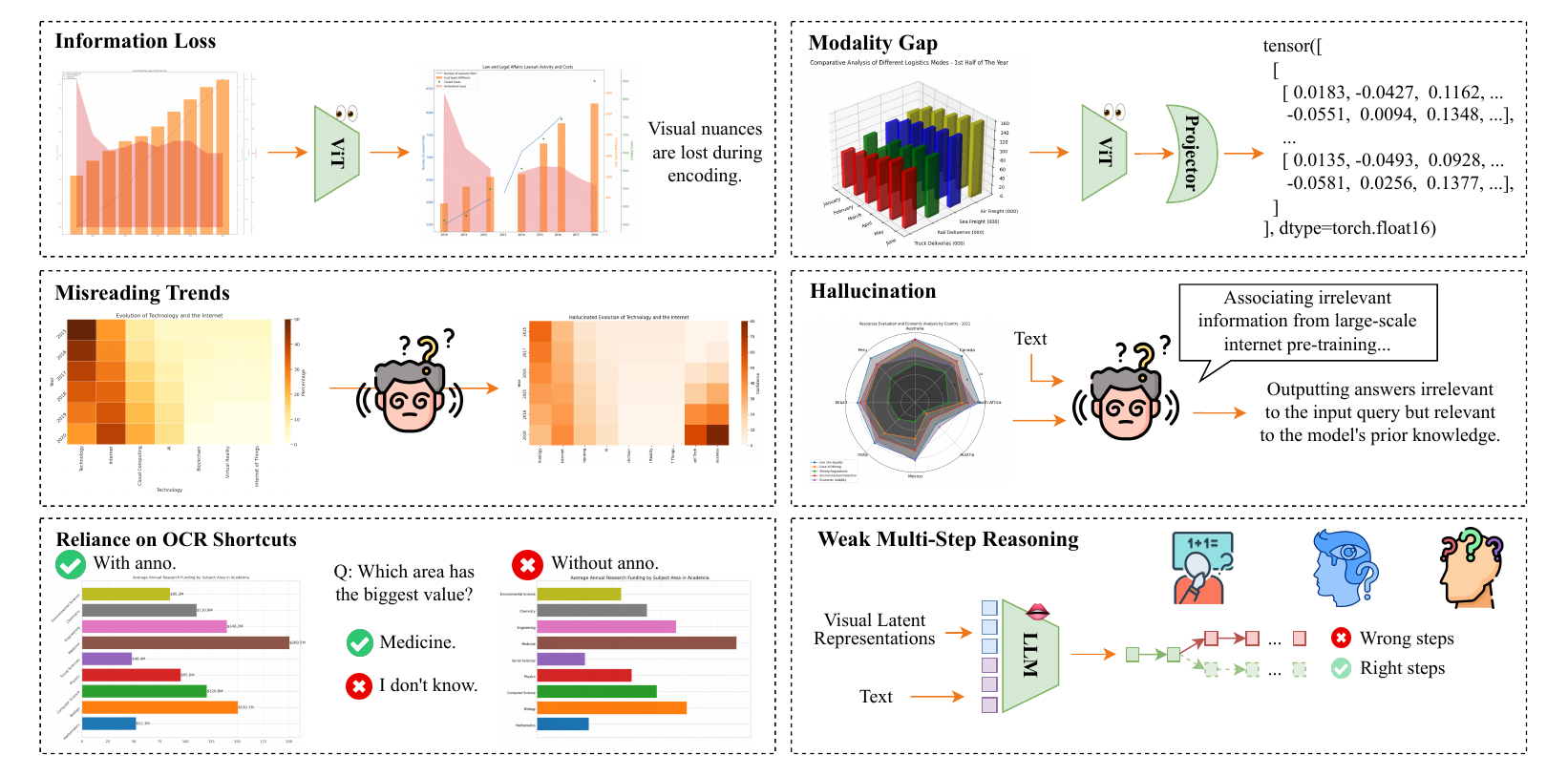}
    \caption{Representative limitations of MLLM-based chart understanding across visual encoding, modality projection, and language decoding. The figure illustrates how information loss and modality gaps can lead to concrete failure behaviors, including trend misreading, hallucination driven by textual priors, reliance on OCR shortcuts, and weak multi-step reasoning.}
    \label{fig: limitations}
\end{figure*}

\subsubsection{Non-Canonical Chart Understanding} 
Historically, Chart Understanding research has predominantly focused on canonical types, such as bar charts, line plots, and histograms, benefiting from their regularized geometry and abundant benchmark data. However, real-world visual communication frequently relies on non-canonical forms, including scientific figures, integrated infographics, maps, flowcharts, and multi-panel layouts. These modalities introduce complex challenges regarding compositional layout, non-linear topology, domain-specific notation, and high-context dependency, necessitating modeling paradigms that transcend simple pattern recognition \cite{reddy2019figurenetdeeplearningmodel, gomezperez2019lookreadenrichlearning, wang2021rlcsdiarepresentationlearningcomputer, lu2021iconqa, tannert2023flowchartqa, dibia2023lidatoolautomaticgeneration, tanaka2023slidevqa, ye2024endtoendvlmsleveragingintermediate, hu2024bliva, arbaz2024genflowchart, baechler2024screenai, omasa2025arrow, huang2025vistruct, he2025flow2codeevaluatinglargelanguage}. Unlike canonical charts, which can often be reduced to data tables or primitive marks, non-canonical charts may require recovering a graph, a document layout, a region map, or a domain-specific semantic structure. Recent advances in this domain can therefore be organized according to the dominant structure that must be fused with language.

\textbf{Graph-Structured and Procedural Visualizations.} Flowcharts, block diagrams, and procedural diagrams encode meaning through node text, arrow direction, branch conditions, and global graph topology. Their answers often depend less on reading an isolated value and more on following a path or executing a decision process. Benchmarks such as FlowchartQA \cite{tannert2023flowchartqa}, FlowVQA \cite{singh2024flowvqamappingmultimodallogic}, FlowLearn \cite{pan2024flowlearnevaluatinglargevisionlanguage}, and Flow2Code \cite{he2025flow2codeevaluatinglargelanguage} reflect this shift from chart perception to visual-logical reasoning, as shown in Fig. \ref{fig: flowchart}. FlowVQA emphasizes information localization, decision-making, and logical progression using human-verified flowcharts, while FlowLearn further provides visual component annotations, OCR, Mermaid code, and VQA pairs for scientific and simulated flowcharts.

This setting has motivated structured intermediate representations and symbolic reasoning pipelines. TextFlow \cite{ye2024endtoendvlmsleveragingintermediate} proposes a two-stage system in which a Vision Textualizer converts flowchart images into text representations, such as Graphviz, Mermaid, or PlantUML, and a Textual Reasoner performs question answering over the recovered representation. This modularity improves controllability and makes it easier to separate visual encoding errors from reasoning errors. Similarly, Arrow-Guided VLM \cite{omasa2025arrow} shows that explicitly detecting nodes, arrow endpoints, and arrow directions before constructing structured prompts can improve flowchart reasoning without fine-tuning. These methods are most suitable when topology or execution logic is central to the task, but their performance depends on faithful recovery of nodes, arrows, OCR text, and graph assembly. Errors in any upstream module can propagate into a plausible but incorrect reasoning path.

\textbf{Scientific Figures and Domain-Specific Visualizations.} Scientific figures often combine statistical plots, equations, domain labels, captions, and multi-panel layouts. As a result, the challenge is not only chart reading but also aligning figure regions with surrounding scientific context and domain-specific terminology. SCICAP \cite{hsu2021scicap, hsu2021scicapgeneratingcaptionsscientific} targets scientific figure captioning, RealCQA \cite{ahmed2023realcqascientificchartquestion} evaluates chart QA over real-world scientific charts with first-order-logic-oriented templates, and SciGraphQA \cite{li2023scigraphqa} constructs multi-turn QA dialogues for academic graphs using paper context. More recent benchmarks such as SCI-CQA \cite{shen2024rethinkingcomprehensivebenchmarkchart} stress authentic scientific literature charts, including multi-plot figures, flowcharts, and structural diagrams. These datasets are suitable for evaluating domain realism and context-aware reasoning, but they also make evaluation harder because visual grounding, OCR, scientific background knowledge, and textual context are tightly entangled.

\textbf{Infographics, Slides, and Multi-Chart Layouts.} Infographics and slide-based visualizations move beyond the isolated chart paradigm. They often combine multiple chart types, icons, pictograms, decorative elements, and explanatory text into a single visual narrative. Early document-oriented datasets such as DocVQA \cite{mathew2021docvqa}, InfographicVQA \cite{mathew2021infographicvqa}, and SlideVQA \cite{tanaka2023slidevqa, tanaka2023slidevqadatasetdocumentvisual} reveal the need for layout reasoning and evidence selection across regions or multiple images. MultiChartQA \cite{zhu2024multichartqa} further focuses on reasoning over multiple co-located charts, while InfoChartQA \cite{lin2025infochartqa} introduces paired infographic and plain charts to isolate the effect of design-driven visual elements, such as icons and pictograms. ChartGalaxy \cite{li2025chartgalaxydatasetinfographicchart} expands this direction by constructing a million-scale dataset for infographic chart understanding and generation from real infographic design patterns. These benchmarks are well suited for testing cross-region synthesis and visual-textual storytelling, but they are difficult to solve with direct image-to-text generation because the model must first decide which region, chart type, or design element is relevant to the question.

\textbf{Spatial and Map-Like Encodings.} Maps, choropleth charts, and geospatial visualizations encode values through regions, shapes, color scales, and spatial adjacency. MapQA \cite{chang2022mapqadatasetquestionanswering} highlights this category by requiring models to answer questions over map visualizations rather than conventional Cartesian plots. Such tasks are suitable for testing region-level grounding and legend-color association, but they expose a different failure mode from canonical charts: an MLLM may recognize the textual legend while failing to bind the correct color, polygon, or geographic region to the requested entity. This makes map understanding a fusion problem involving visual segmentation, spatial priors, and textual grounding.

\textbf{Data-Centric Innovation and Synthetic Curation.} Progress in non-canonical modeling is intrinsically linked to instruction-tuning data, because many non-canonical chart types are underrepresented in standard chart corpora. Recent datasets have shifted from simple QA pairs to complex, multi-turn reasoning and structurally grounded supervision \cite{lee2016viziometricsanalyzingvisualinformation, li2023scigraphqa, ahmed2023realcqascientificchartquestion, zhu2024multichartqa, zhang2024multimodal, zhang2024first, yue2024mmmu, singh2024flowvqamappingmultimodallogic, qiu2024genchar, pan2024flowlearnevaluatinglargevisionlanguage, li2024multimodal, li2024multimodalarxivdatasetimproving, yang2025scaling, lin2025infochartqa, li2025chartgalaxydatasetinfographicchart}. Three primary curation strategies have emerged. 

\begin{itemize}
    \item \textbf{Conversation-Driven Annotation} leverages MLLMs to simulate dialogue. SciGraphQA \cite{li2023scigraphqa}, derived from SciCap+ \cite{yang2023scicapknowledgeaugmenteddataset}, utilizes multi-turn interactions to embed contextually linked reasoning steps, while Multimodal ArXiv \cite{li2024multimodal} employs a crawl-and-query method to harvest scientific contents from arXiv. This strategy is suitable for modeling interactive analytical workflows, but the generated conversations may reflect teacher-model biases or superficial dialogue patterns.
    
    \item \textbf{Human-Verified Curation} ensures domain fidelity. MultiChartQA \cite{zhu2024multichartqa} and MMMU \cite{yue2024mmmu} rely on human experts and university students, respectively, to craft high-quality QA pairs from authentic literature, prioritizing correctness over scale. This strategy is most suitable for high-stakes or domain-specific evaluation, but it is expensive and difficult to scale across diverse non-canonical chart layouts.

   \item \textbf{Code-Driven Synthesis} addresses the data scarcity of complex logic. Frameworks like Multimodal Self-Instruct \cite{zhang2024multimodal, zhang2024multimodalselfinstructsyntheticabstract} and CoSyn \cite{yang2025scaling} employ imagination-to-code pipelines, where LLMs generate scenarios and render them via visualization libraries (e.g., Matplotlib, Vega-Lite, Graphviz). This allows for the generation of ground-truth structural annotations and Chain-of-Thought (CoT) reasoning paths. CoSyn even introduces varying ``personas'' to ensure stylistic and conceptual diversity across examples. Hybrid approaches, such as FlowVQA \cite{singh2024flowvqamappingmultimodallogic} and FlowCE \cite{zhang2024first}, combine web-mining with synthetic augmentation by using the keyword ``flowchart'' from Google, Baidu, wikiHow, Instructables blogs, and FloCo repositories, while InfoChartQA \cite{lin2025infochartqa} and ChartGalaxy \cite{li2025chartgalaxydatasetinfographicchart} focus on the diversity of infographic styles, balancing real-world visual noise with synthetic structural clarity. This strategy scales well and provides controllable ground truth, but synthetic layouts may underrepresent the ambiguity, noise, and design irregularity of real-world non-canonical charts.
\end{itemize}

Together, these efforts represent a paradigm shift in data curation for non-canonical Chart Understanding, from static, isolated instances to rich, diverse, and dynamic multimodal scenarios that better align with the reasoning capabilities expected of modern MLLMs.

\paragraph{Insight.} Non-canonical charts expose the limits of models trained primarily on regular statistical plots. Unlike bar charts or line charts, flowcharts, infographics, maps, scientific figures, and multi-chart layouts require topology recovery, layout reasoning, cross-region synthesis, and domain-specific knowledge. Successful approaches therefore tend to rely on structured intermediate representations, symbolic execution, region-level grounding, or hybrid pipelines rather than direct image-to-text generation alone. This suggests that non-canonical chart understanding is not merely a data scaling problem, but a deeper fusion problem involving visual structure, textual semantics, spatial relations, and logical reasoning. Progress in this direction is essential for closing the gap between benchmark-centered chart understanding and real-world visual information analysis.

\section{Limitations} \label{sec:limitations}
Despite the rapid proliferation of Multimodal Large Language Models (MLLMs), their application to chart understanding remains constrained by fundamental architectural and cognitive deficits. These limitations can be categorized into perceptual encoding failures, where visual fidelity is lost, and cognitive alignment gaps, where visual signals fail to ground symbolic reasoning as shown in Fig. \ref{fig: limitations}. The following discussion connects these architectural bottlenecks with observable failure behaviors, such as axis-value misreading, legend-series confusion, OCR shortcut reliance, and visually unsupported explanations.

\subsection{Perceptual Encoding and Granularity Deficits}
A primary bottleneck in current MLLM frameworks is the inability of Vision Encoder to preserve the high-frequency spatial details required for accurate chart interpretation \cite{kamath2023whatsupvisionlanguagemodels, tong2024eyeswideshutexploring, kamoi2025visonlyqalargevisionlanguage}. Current MLLMs fail to capture the rich, multi-layered details inherent in these complex visual representations. Unlike natural images, where semantic meaning is often invariant to minor resolution drops, chart images rely on precise geometric primitives (e.g., grid lines, tick marks, exact marker positions) to convey quantitative information. Current encoders, predominantly optimized for broad semantic alignment on natural image-text pairs, frequently exhibit "geometric blindness," failing to encode the subtle spatial hierarchies and nuanced graphical variables (e.g., stroke width, color gradients) essential for distinguishing data series \cite{liu2025perception}. This results in a loss of informational fidelity that effectively renders the model blind to the chart's structural syntax. In practice, this bottleneck appears as axis-value misalignment: a model may assign a bar, point, or line segment to the wrong tick value, especially under dense ticks, small markers, non-zero baselines, or low-resolution rendering. Such errors are not merely numerical slips; they reveal that continuous chart geometry is not preserved strongly enough for downstream symbolic reasoning.

Consequently, MLLMs often resort to an "OCR shortcut," bypassing genuine visual reasoning in favor of textual extraction. As evidenced by recent benchmarks \cite{xu2023chartbench, islam2024largevisionlanguagemodels}, performance degrades precipitously when explicit alphanumeric annotations are removed. Models demonstrate a heavy reliance on legends, titles, and data labels to hallucinate trends, rather than perceiving the visual slope or relative magnitude of elements. For instance, a model may correctly transcribe a label "Revenue: \$5M" but fail to determine if the associated bar is the tallest in the series without text. A related failure is legend-series confusion: the model may read the legend labels correctly but bind a color or marker to the wrong data series, producing an answer that is textually fluent but visually mismatched. This dependency suggests that current systems perform linguistic pattern matching over extracted text rather than grounding their output in visual evidence \cite{islam2024large}. 

Moreover, the intelligence of these MLLMs in interpreting charts often stems not from a deep visual comprehension, but from their vast knowledge repositories amassed during large-scale internet pre-training \cite{mukhopadhyay2024unraveling, hou2024vision, guo2024understandinggraphicalperceptiondata}. When presented with a chart, an MLLM might leverage its textual understanding of common chart types, axes labels, or domain-specific terminology to make plausible inferences. For example, if a chart is labeled "Global Temperature Trends", the model's pre-existing knowledge about climate change and common graphical representations of temperature might lead it to accurate conclusions, even if Vision Encoder has only superficially processed the visual elements.

This superficial encoding further manifests as a lack of robustness to visual perturbation. Minor stylistic alterations, such as changes in aspect ratio, color palettes, or font styles, can induce catastrophic performance drops, indicating that the visual representations are not invariant to style \cite{mukhopadhyay2024unraveling}. Furthermore, the generalization gap remains significant as encoders trained on standard bar or line charts frequently falter on topologically distinct visualizations like radar plots or tree diagrams, highlighting a failure to learn a generalized grammar of graphics \cite{zhang2025understandinggraphicalperceptionlarge}.

\subsection{Cognitive Alignment and Fusion-Induced Hallucination}
Even when visual features are successfully encoded, a critical modality gap persists between the visual representation and the Language Decoder's symbolic processing capabilities. The core challenge lies in the Language Decoder's inability to perform visual arithmetic, the translation of visual magnitudes into precise numerical reasoning \cite{huang2025visionlanguagemodelsstruggle}.

While Vision Encoder may capture the geometry of a trend, the decoder often fails to translate this continuous signal into discrete logical statements. This misalignment leads to imprecise or qualitative outputs (e.g., stating "sales increased" rather than "sales rose by approximately 20\%") and indicates a disconnect between visual perception and linguistic articulation \cite{liu2025perception}. The model struggles to map the continuous manifold of visual embeddings onto the discrete symbolic space required for mathematical logic, resulting in an inability to perform comparative inference or aggregation without explicit textual cues. Observable errors include comparing the wrong bars after a series-binding error, averaging visually misread values, or generating a correct-looking arithmetic chain from incorrect visual operands.

This disconnect is a primary driver of fusion-induced hallucination in chart analysis. From an information-fusion perspective, chart hallucination can be viewed as a failure of evidence-conditioned generation. Let \(I\) denote a chart image, \(q\) denote the user question, \(E_v(I)\) denote the visual tokens produced by Vision Encoder, and \(y\) denote the generated answer. Ideally, the decoder should estimate \(p(y \mid E_v(I), q)\), where the answer is strongly conditioned on chart-specific visual evidence. In practice, however, dense graphical information is compressed by Vision Encoder and the visual projector before reaching Language Decoder, while the decoder retains strong linguistic priors learned from large-scale textual pre-training. When \(E_v(I)\) is incomplete, ambiguous, or weakly aligned with textual tokens, generation may be dominated by \(p(y \mid q)\) or topic-level priors rather than by visual evidence.

Unlike hallucinations in general NLP, which often stem from training data bias, chart hallucinations therefore frequently appear as visually ungrounded yet linguistically plausible claims \cite{islam2024large, hou2024vision}. For example, a model might fabricate a "seasonal dip" in a financial chart simply because its pre-training corpus associates finance discussions with seasonality, even if the visual data depicts a monotonic increase. Similarly, it may describe a non-existent correlation after correctly reading the chart title but failing to align data points across axes. This phenomenon undermines the trustworthiness of MLLMs in critical decision-support scenarios \cite{guo2024understandinggraphicalperceptiondata}.

Finally, higher-order reasoning remains a systemic weakness. MLLMs struggle with tasks requiring multi-hop logic, such as causality inference ("Why did sales drop?") or predictive modeling ("What is the projected value for Q4?"). Current architectures favor descriptive captioning over analytical deduction, often failing to synthesize auxiliary context with visual data to generate actionable insights. In non-canonical charts, this weakness further appears as layout and topology failure: the model may follow the wrong panel, arrow, node, or spatial path before constructing a plausible reasoning trace. This limitation confirms that while MLLMs possess vast encyclopedic knowledge, they lack the "System 2" reasoning capabilities necessary to function as reliable data analysts.

\section{Promising Solutions} \label{sec:future_directions}
To overcome the architectural and cognitive bottlenecks inherent in current MLLMs, recent research has pivoted towards two synergistic frontiers: enhancing the fidelity of visual-semantic alignment and incentivizing robust reasoning through advanced post-training paradigms.

\subsection{Refining Perceptual Fidelity and Robustness}
The foundational limitation of general-purpose vision encoders (e.g., CLIP) lies in their coarse-grained semantic alignment, which often discards the high-frequency structural details required for chart interpretation. Addressing this requires a shift from general pre-training to domain-adaptive refinement.

\paragraph{Discriminative Fine-Tuning with Hard Negatives.} Standard fine-tuning often fails to correct "bag-of-words" behavior in Vision Encoder \cite{yuksekgonul2023visionlanguagemodelsbehavelike}. A critical advancement involves the integration of hard negative mining into contrastive learning objectives \cite{liu2025perception}. By augmenting training data with "hard negative" captions, descriptions that are semantically plausible but factually contradictory (e.g., describing a monotonic increase as a plateau), models are forced to learn discriminative features rather than relying on superficial correlations \cite{yuksekgonul2023visionlanguagemodelsbehavelike}. This methodology compels the encoder to attend to precise visual evidence, such as trend directionality and marker distinctiveness, significantly reducing information loss during encoding as shown in Tab. \ref{tab: hard negative}.

\renewcommand{\arraystretch}{1.5}
\begin{table*}
    \centering
    \begin{tabular}{ll|cccccc}
        \hline
        Method & Average & FigureQA & DVQA-E & DVQA-H & PlotQA & ChartQA & ChartBench \\
        \hline
         Original CLIP & 25.5 & 50.0 & 25.8 & 25.6 & 8.9 & 12.8 & 4.8 \\
         
         + Supervised Fine-Tuning & 41.5 \textcolor{mygreen}{\tiny +16.0} & 48.6 & 28.9 &  27.2 & 22.1 & 18.8 & 7.4 \\
         
         + Hard Negatives & 51.4 \textcolor{mygreen}{\tiny +25.9} & 82.0 & 65.2  & 61.0 & 54.1 & 29.7 & 16.2 \\
         \hline
    \end{tabular}
    \caption{Image-to-text retrieval evaluation accuracy on original CLIP and fine-tuned CLIPs. The experiment results suggest that fine-tuning CLIP with chart-specific pairs and their hard negative captions can effectively enhance CLIP's performance.}
    \label{tab: hard negative}
\end{table*}

\begin{figure*}
    \centering
    \includegraphics[width=\linewidth]{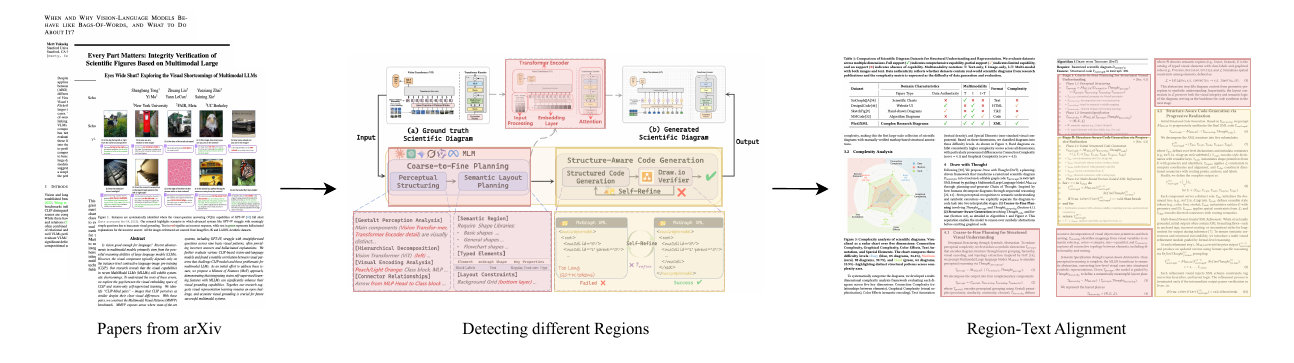}
    \caption{An illustration of Fine-Grained Visual Grounding. Instead of mapping the entire chart image to a caption, recent research aligns discrete visual regions with their corresponding annotations.}
    \label{fig: region}
\end{figure*}

\paragraph{Fine-Grained Visual Grounding.} To mitigate the model's reliance on OCR-derived text, recent frameworks emphasize region-level alignment \cite{shi2024mattersintegrityverificationscientific}. As shown in \ref{fig: region}, rather than mapping a global image embedding to a caption, these approaches explicitly align discrete visual regions (e.g., a specific bar or axis tick) with their corresponding textual annotations. This pixel-to-value grounding transforms the encoding process from a general scene understanding task into a structured extraction task, ensuring that numerical values are topologically linked to their visual representations.

\paragraph{Invariance through Dynamic Resolution and Synthesis.} Addressing the lack of robustness requires both architectural and data-centric interventions. Architecturally, dynamic resolution modules have emerged as a solution to the aspect-ratio distortion problem, allowing encoders to process charts in their native geometries without destructive resizing. Concurrently, data diversity is being addressed through large-scale synthetic augmentation, which generates vast permutations of chart styles, taxonomies, and noise levels \cite{mukhopadhyay2024unraveling, zhang2025understandinggraphicalperceptionlarge}. By training on this diverse "visual hull," MLLMs learn to extract invariant structural properties rather than overfitting to specific aesthetic conventions.

\subsection{Incentivizing Cognitive Reasoning and Verification}
While perceptual enhancements ensure accurate data encoding, transforming that data into actionable insight requires a paradigm shift in how models are trained to reason.

\paragraph{Reinforcement Learning for Chain-of-Thought (CoT).} A transformative direction in MLLM training is the transition from Supervised Fine-Tuning (SFT) to Reinforcement Learning (RL) for reasoning tasks. As exemplified by the recent DeepSeek-R1 architecture \cite{deepseekai2025deepseekr1incentivizingreasoningcapability}, RL can be employed to incentivize the generation of explicit thinking processes rather than just final answers. Unlike SFT, which mimics human labels, RL allows the model to explore reasoning paths within a simulated environment, receiving reward signals based on the logical validity and factual accuracy of its deductions. This approach has been shown to induce emergent "System 2" thinking \cite{li202512surveyreasoning}, enabling models to self-correct and develop complex analytical strategies, such as verifying a trend against a legend before concluding, thereby significantly surpassing the reasoning depth of standard pre-trained models.

\begin{figure*}
    \centering
    \includegraphics[width=\linewidth]{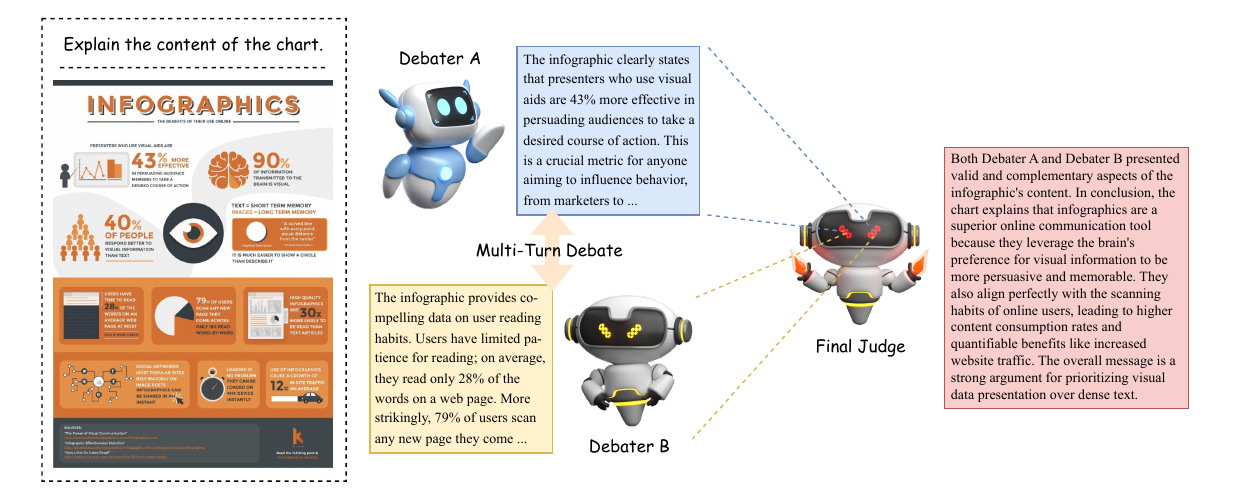}
    \caption{An illustration of Multi-Agent Debate.}
    \label{fig: debate}
\end{figure*}

\paragraph{Multi-Agent Debate and Collaborative Refinement.} To address the stochastic nature of hallucination, the Multi-Agent Debate framework in Fig. \ref{fig: debate} offers a robust inference-time solution \cite{kim2024can}. This paradigm replaces the single-model monologue with a collaborative dialectic, where diverse agents (or divergent personas within a single model) critique and refine each other's interpretations. For instance, a skeptic agent may challenge an initial finding by demanding visual evidence, prompting a verifier agent to cross-reference the chart data. This adversarial collaboration acts as a cognitive filter, pruning hallucinated insights and converging on a consensus that is statistically more likely to be factually grounded. Such mechanisms mirror human peer-review processes, significantly enhancing trust and reliability in high-stakes analysis.

\subsection{Deployment-Oriented Chart Intelligence in Analytical Workflows}
Beyond benchmark performance, practical chart understanding systems should be integrated into analytical workflows rather than deployed as isolated question-answering modules. A deployment-oriented pipeline typically begins with chart or document ingestion and chart-region detection, proceeds through chart-type and layout analysis, performs visual-textual grounding, and then selects a task-specific reasoning path such as factoid retrieval, data extraction, chart-to-code generation, or open-ended explanation. In scientific reading, financial reporting, business intelligence dashboards, and public-policy analysis, the system should not only return chart-related answers, but also expose the visual evidence, extracted values, and intermediate representations that support them.

Such deployment requires adaptive computation because the same pipeline need not invoke the most expensive model or reasoning strategy for every query. For clean canonical charts and simple factoid questions, lightweight OCR, derendering, or smaller chart-specific VLMs may be sufficient. For numerical analysis, chart-to-table translation, chart-to-code generation, or structural reconstruction, intermediate representations such as tables, executable code, SVG/XML, or dynamically selected tools provide stronger evidence and verification. For high-uncertainty, high-stakes, or open-ended analytical tasks, costlier mechanisms such as multi-agent debate, RL-optimized tool selection, or self-verification can be triggered selectively. Caching recovered tables, code, or layout representations for repeated dashboards and reports can further reduce latency while preserving traceability.

Human-in-the-loop interaction is also essential for trustworthy deployment. Instead of returning only a fluent textual answer, deployed systems should provide interpretable artifacts, such as highlighted chart regions, recovered data tables, executable analysis scripts, confidence estimates, or warnings when the answer depends on weak OCR, low-resolution visual evidence, or ambiguous layout parsing. Human analysts can then inspect, correct, or approve intermediate outputs before they are used in downstream decisions, and these corrections can in turn provide audit trails or preference signals for later model improvement. From an information-fusion perspective, real-world deployment therefore shifts the goal from end-to-end answer generation to workflow-aware fusion of visual evidence, symbolic representations, external tool outputs, and human expertise.

\section{Conclusion}
In this survey, we have presented a comprehensive analysis of chart understanding through the lens of multimodal information fusion. We systematically charted the evolution from classical computer vision techniques to the sophisticated fusion paradigms employed by modern Multimodal Large Language Models (MLLMs). By structuring the domain's challenges, tasks, and datasets—notably introducing a distinction between canonical and non-canonical charts—we have provided a clear roadmap of the current landscape. Our detailed taxonomy of MLLM-based methods reveals a clear trend towards more complex fusion strategies, including the use of intermediate symbolic representations and dynamic, expert-based models.

However, our analysis also identified critical limitations in current MLLMs, namely deficits in perceptual fidelity and cognitive reasoning, which lead to a lack of robustness and a tendency for ungrounded hallucination. To address these gaps, we highlighted promising future directions centered on enhancing visual-semantic alignment, incentivizing robust reasoning through advanced training paradigms like reinforcement learning and multi-agent debate, and integrating chart intelligence into deployment-aware analytical workflows. Ultimately, this survey serves not only as a structured review but also as a call to action: for the field to move beyond superficial integration and toward a new generation of MLLMs capable of true, reliable, and verifiable visual-linguistic information fusion for data intelligence.

\section{Acknowledgements}
This work is supported by the National Science Foundation of China under Grant 62506249, the National Major Scientific Instruments and Equipments Development Project of National Natural Science Foundation of China under Grant 62427820, the Natural Science Foundation of Sichuan under grant 2024NSFSC1462, and the Fundamental Research Funds for the Central Universities under grant YJ202342.

\printcredits

\bibliographystyle{elsarticle-num}

\bibliography{cas-refs}



\end{document}